%% file: harpa_arxiv.tex
\documentclass{article} % For LaTeX2e
\usepackage{arxiv,times}
\input{math_commands.tex}

\usepackage{hyperref}
\usepackage{url}

\usepackage{colortbl}

\usepackage[frozencache,cachedir=.]{minted}
\setminted{style=default}
\usepackage{listings}

\usepackage{microtype}
\usepackage{pdfpages}
\usepackage{graphicx}
\usepackage{comment}
\usepackage{wrapfig} 
\usepackage{inconsolata}
\usepackage{xspace} 
\usepackage{listings}
\usepackage{makecell}
\usepackage{array}
\usepackage{subcaption}
\usepackage{multirow}
\usepackage{comment}
\usepackage{booktabs}
\usepackage{multirow}
\usepackage{amsmath}
\usepackage{wrapfig}
\usepackage{breqn}
\usepackage{caption} % for \captionof{}
\usepackage[most]{tcolorbox}
\usepackage{mdframed}
\usepackage{pifont}
\usepackage{adjustbox}
\usepackage{fontawesome5}
\usepackage{longtable}
\newcommand{\showtracebox}[4]{%
\begin{tcolorbox}[
    colback=gray!5,
    colframe=blue!65!black,
    coltitle=white,
    enhanced,
    breakable,
    listing only,
    title=#2,
    rounded corners,
    boxrule=0.7mm,
    width=\textwidth,
    arc=2mm
]
\inputminted[
    fontsize=\small,
    breaklines=true,
    breaksymbolleft=\texttt{...},
    linenos=false
]{xml}{#3}
\end{tcolorbox}
\vspace{-0.5em}
\captionof{listing}{#4}\label{#1}%
}

\newcommand{\showjsonbox}[5]{%
\begin{tcolorbox}[
    colback=teal!5,          
    colframe=teal!60!black,
    coltitle=white,
    enhanced,
    breakable,
    listing only,
    title=#2,
    rounded corners,
    boxrule=0.7mm,
    width=\textwidth,
    arc=3mm
]
\inputminted[
    fontsize=\small,
    breaklines=true,
    breaksymbolleft=\texttt{...},
    linenos=false,
    autogobble
]{#4}{#3}
\end{tcolorbox}
\vspace{-0.5em}
\captionof{listing}{#5}\label{#1}%
}

\newcommand{\showmintedbox}[4]{%
\begin{tcolorbox}[
    colback=pink!5,
    colframe=pink!50!black,
    coltitle=white,
    enhanced,
    breakable,
    listing only,
    title=#2,
    rounded corners,
    boxrule=0.7mm,
    width=\textwidth,
    arc=4mm
]
\inputminted[
    fontsize=\small,
    breaklines=true,
    breaksymbolleft=\texttt{...},
    linenos=false,
    escapeinside=||,
]{text}{#3}
\end{tcolorbox}
\vspace{-0.5em}
\captionof{listing}{#4}\label{#1}%
}

\newcommand{\showscorerbox}[4]{%
\begin{tcolorbox}[
    colback=purple!5,
    colframe=purple!50!black,
    coltitle=white,
    enhanced,
    breakable,
    listing only,
    title=#2,
    rounded corners,
    boxrule=0.7mm,
    width=\textwidth,
    arc=4mm
]
\inputminted[
    fontsize=\small,
    breaklines=true,
    breaksymbolleft=\texttt{...},
    linenos=false,
    escapeinside=||,
]{text}{#3}
\end{tcolorbox}
\vspace{-0.5em}
\captionof{listing}{#4}\label{#1}%
}

\newcommand{\eat}[1]{}

\title{%
  \raisebox{-0.2em}{\includegraphics[height=1.5em]{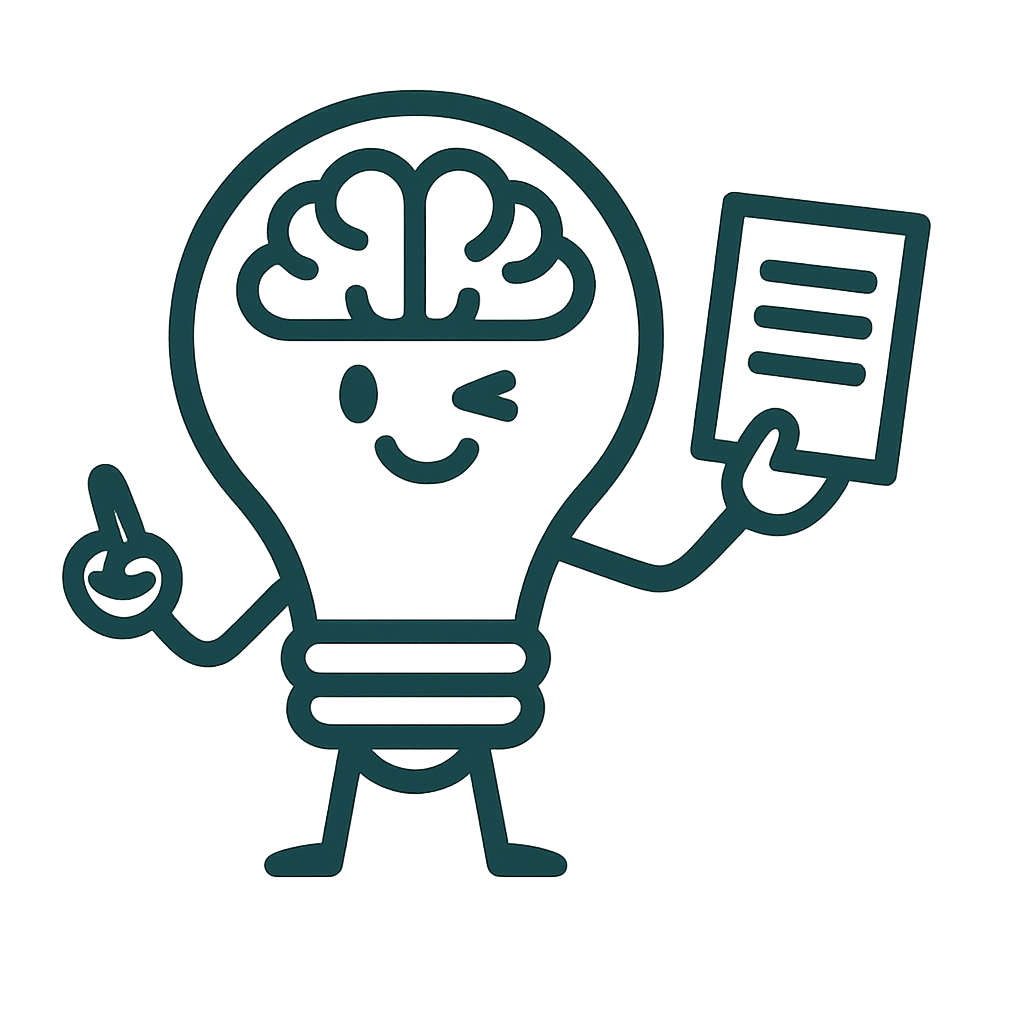}}%
   HARPA: A Testability-Driven, Literature-Grounded Framework for Research Ideation%
}
\author{
\textbf{Rosni Vasu\textsuperscript{\rm $\ddagger$\thanks{Work performed while at Ai2}}}, 
\textbf{Peter Jansen\textsuperscript{\rm $\#$,$\spadesuit$}}, 
\textbf{Pao Siangliulue\textsuperscript{\rm $\spadesuit$}},
\textbf{Cristina Sarasua\textsuperscript{\rm $\ddagger$}},\\ \vspace{0.5em}
\textbf{Abraham Bernstein\textsuperscript{\rm $\ddagger$}},
\textbf{Peter Clark\textsuperscript{\rm $\spadesuit$}}, 
\textbf{Bhavana Dalvi Mishra\textsuperscript{\rm $\spadesuit$}} \\
\textsuperscript{$\ddagger$}University of Zurich \quad
\textsuperscript{$\#$}University of Arizona \quad
\textsuperscript{$\spadesuit$}Allen Institute for AI \\
\vspace{0.5em}
\texttt{rosni@ifi.uzh.ch, bhavanad@allenai.org}
}

\iclrfinalcopy % Uncomment for camera-ready version, but NOT for submission.

\begin{document}

\maketitle

\begin{abstract}
While there has been a surge of interest in automated scientific discovery (ASD), especially with the emergence of LLMs, it remains challenging for tools to generate hypotheses that are both testable and grounded in the scientific literature. Additionally, 
existing ideation tools are not adaptive to prior experimental outcomes.
We developed HARPA%\footnote{\scriptsize All code and data used in this paper will be made publicly available at GitHub Link: (removed for review).} 
to address these challenges by incorporating the ideation workflow inspired by human researchers. HARPA first identifies emerging research trends through literature mining, then explores hypothesis design spaces, and finally converges on precise, testable hypotheses by pinpointing research gaps and justifying design choices.
Our evaluations show that HARPA-generated hypothesis-driven research proposals perform comparably to a strong baseline AI-researcher across most qualitative dimensions (e.g., specificity, novelty, overall quality), but achieve significant gains in feasibility(+0.78, p$<0.05$, bootstrap) and groundedness (+0.85, p$<0.01$, bootstrap) on a 10-point Likert scale.
When tested with the ASD agent (CodeScientist), HARPA produced more successful executions (20 vs. 11 out of 40) and fewer failures (16 vs. 21 out of 40), showing that expert feasibility judgments track with actual execution success.
Furthermore, to simulate how researchers continuously refine their understanding of what hypotheses are both testable and potentially interesting from experience, HARPA learns a reward model that scores new hypotheses based on prior experimental outcomes, achieving approx. a 28\% absolute gain over HARPA's untrained baseline scorer. Together, these methods represent a step forward in the field of AI-driven scientific discovery.
\end{abstract}

%%%%%%%%%%%%%%%%%%%

\input{chapters/introduction}
\input{chapters/related_work}

\input{chapters/method_harpa}

\input{chapters/method_harpa_scorer}

\input{chapters/experiments}

\input{chapters/results}

\input{chapters/conclusion}

%%%%%%%%%%%%%%%%%

%uncomment later
\eat{
\subsubsection*{Author Contributions}
If you'd like to, you may include  a section for author contributions as is done
in many journals. This is optional and at the discretion of the authors.

\subsubsection*{Acknowledgments}
Use unnumbered third level headings for the acknowledgments. All
acknowledgments, including those to funding agencies, go at the end of the paper.
}

\subsubsection*{Ethics Statement}
We honor the Code of Ethics. No personally identifiable information is used in this work. The human evaluators were hired from Upwork using a detailed job post. We had Institutional Review Board (IRB) approval for obtaining written consent from our human evaluators. We shared an example task sheet with complete instructions during the recruitment. The evaluators were duly compensated based on minimum wage in the respective countries and always above their quotation.

\bibliography{iclr2026_conference}
\bibliographystyle{iclr2026_conference}

\appendix
\section{Appendix}
\input{appendix/human_eval}
\input{appendix/harpa_scorer}
%%% add all the appendix files here
\input{appendix/harpa_stages}

\end{document}

%% file: math_commands.tex
\usepackage{amsmath,amsfonts,bm}

\def\eqref#1{equation~\ref{#1}}
\def\1{\bm{1}}

\DeclareMathAlphabet{\mathsfit}{\encodingdefault}{\sfdefault}{m}{sl}
\SetMathAlphabet{\mathsfit}{bold}{\encodingdefault}{\sfdefault}{bx}{n}

\newcommand{\yes}{\checkmark\xspace}

\DeclareRobustCommand{\somewhat}{\raisebox{-0.2ex}{$\sim$}\xspace}
\newcommand{\no}{$\times$\xspace}

\newenvironment{ite}{                     % list without par spacings
     \parskip 0cm \begin{itemize} \parskip 0cm \parsep 0cm \itemsep 0cm \topsep 0cm}{
        \end{itemize}} %  \parskip 0cm}

%% file: chapters/introduction.tex
\section{Introduction}\label{sec:intro}

Scientific discovery fundamentally depends on effective hypothesis generation—a creative, iterative, and cognitively complex process. In the past year, advances in large language models (LLMs) have revitalized the field of Automated Scientific Discovery (ASD) and AI-assisted ideation, by providing the foundations for agents that can autonomously execute experiments~\citep{lu2024ai,gottweis2025towards,jansen2025codescientist,li2024mlrcopilotautonomousmachinelearning}. At the same time, these models have been applied to generate novel research ideas~\citep{radensky2024scideator,pu2024ideasynth,baek2024researchagent, wang2023scimon,li2024chain}, supplying candidate ideas for the experimental agents to explore.

One of the central challenges of automated scientific discovery is that the hypotheses generated by large language models rarely rise to the level of breakthrough discoveries~\citep{gottweis2025towards}. While such hypotheses may be novel or creative, they are frequently infeasible as research proposals~\citep{si2025ideation}. Common issues include limited grounding in literature, omission of critical methodological details, and reliance on resource-intensive experimental designs that exceed the capacity of ASD agents. These challenges mirror findings from prior studies, where ideation systems often produce ideas that are too abstract to be actionable, require substantial human intervention to refine into testable research proposals~\citep{li2024chain,vasu2025hyper, radensky2024scideator,wang2023scimon,pu2024ideasynth}, or lack mechanisms to balance novelty with feasibility~\citep{li2024mlrcopilotautonomousmachinelearning,jansen2025codescientist,gottweis2025towards}.

\begin{figure}[t]
    \centering
    \includegraphics[width=0.9\linewidth]{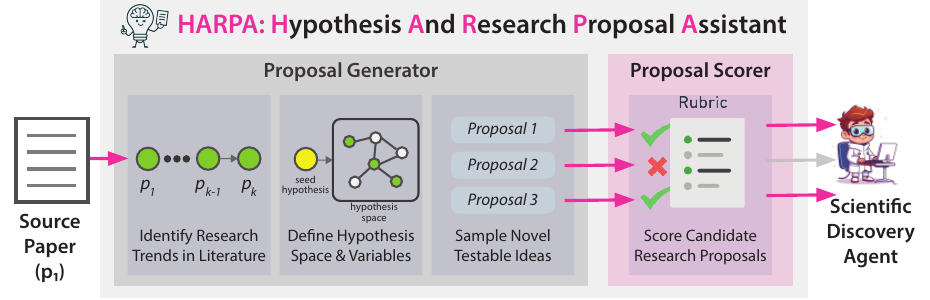}
    \caption{Overview of HARPA. Starting from a source paper, seed hypothesis derived from literature trends, HARPA constructs a \textit{world model} of variables, values, and supporting evidence. The proposal generator consists of three stages (trend identification, hypothesis space exploration for divergence, proposal sampling for convergence) to produce candidate hypothesis-driven research proposals. A dedicated scorer employs reasoning-based reward model based on prior execution evidences to evaluate testability w.r.t target ASD agent.
    }
    \label{fig:harpa-overview}
\end{figure}

In this work, we present HARPA --- \textbf{H}ypothesis \& \textbf{R}esearch \textbf{P}roposal \textbf{A}ssistant --- a novel multi-stage computational framework that generates literature-grounded research proposals with specific hypotheses well-supported for ASD systems. HARPA is composed of a \emph{proposal generator} and a \emph{scorer}, as shown in Figure~\ref{fig:harpa-overview}. HARPA generates significantly more feasible research ideas by virtue of its generation approach being strongly grounded in the scientific literature: first identifying relevant research trends related to a user's hypothesis of interest, then systematically exploring the constructed hypothesis space of variables and their values, and finally converging on novel (and significantly more feasible) hypotheses as well-reasoned combinations of variables and research trends that fill identifiable research gaps in the literature.

We evaluate HARPA using a combination of expert human studies as well as ASD agents. We empirically show that HARPA-generated proposals are more feasible and better grounded in the scientific literature than those produced by contemporary systems. Beyond these gains, we further demonstrate that HARPA's reward-trained scorer, distilled in RM-R1 style~\citep{chen2025rm}, can predict which research proposal is most likely to execute on the ASD agent. Unlike a black-box classifier, the scorer produces rubric-style reasoning traces, interpretable justifications generated from the proposal content and conditioned on the ASD agent's capabilities, trained to reflect patterns distilled from prior execution outcomes. This enables HARPA to incorporate feedback from prior experimental evidences to selectively generate proposals tailored to the strength and constraints of specific ASD agent - much as a professor might guide a student toward research ideas aligned with the student's prior knowledge and expertise.

Our results empirically demonstrate that  HARPA nearly doubles the scientific output of automated discovery systems, measured as the number of successfully executed experiments, while also reducing costs by pruning infeasible proposals unlikely to succeed before they are attempted. Our contributions:
\begin{ite}
\item \textbf{HARPA:} a novel literature-grounded framework for hypothesis generation for ASD systems, that combines identifying research trends, hypothesis space construction, and testability-aware convergence to generate proposals that are novel and executable.
\item \textbf{Empirical demonstration:} studies with both human domain experts and automated scientific discovery systems showing that HARPA-generated proposals are rated higher in feasibility (+0.78, p$\leq$0.05) and literature-grounding (+0.85, p$\leq$0.01), and achieve higher execution success (\somewhat80\% more, 20 vs. 11), compared to competing systems.
\item \textbf{Learned feasibility:} We demonstrate that HARPA's scorer --- an interpretable reward model distilled from actual execution traces, can predict which research proposals are most likely to be executable by a given ASD agent, significantly saving time and cost by selectively pruning hypotheses that the system is unlikely to execute. HARPA's scorer achieves a +0.28 absolute, 53\% relative gain over the untrained baseline scorer.
\item A publicly available implementation of this approach (HARPA), including the HARPA-Scorer model (to be released on Hugging Face), and first large-scale ASD execution traces and preference dataset to support reproducibility and future research.
\end{ite}

Together, these contributions represent a step toward more capable hypothesis generation tools and help advance the rapidly growing field of AI-driven scientific discovery.

%% file: chapters/related_work.tex
\section{Related Work \label{sec:related-work}}

\paragraph{Human hypothesis generation.} Cognitive science highlights that scientific hypothesis generation is a complex iterative process involving strategies such as analogical reasoning and model based thinking, where simplified representations guide inquiry~\citep{dunbar2000scientists,nersessian2010creating,klahr1999studies}. The Scientific Discovery as Dual Search (SDDS) model~\citep{klahr1988dual} identifies strategies such as searching memory for relevant hypotheses and generalizing from experimental results, underscores the need for the ASD systems that can reason over structured hypothesis spaces and adapt from experimental feedback. Prior work has also examined how researchers navigate the broader scientific landscape, where scientists often favor incremental, topic-adjacent experiments~\citep{rzhetsky2015choosing}, with only a minority pursuing riskier but higher-impact directions~\citep{foster2015tradition}. 

\begin{wraptable}{r}{0.65\textwidth} % right side, half-width
\centering
\small
\renewcommand{\arraystretch}{1.1}
\adjustbox{max width=\linewidth}{
\begin{tabular}{lccccc}
\toprule
\textbf{System} & 
\makecell{1) Grounded \\ ideas?} & 
\makecell{2) Domain- \\ General?} & 
\makecell{3) Full \\ proposal?} & 
\makecell{4) ASD \\ Feasibility?} & 
\makecell{5) Adaptive?} \\
\midrule
GPT-5                 & \no  & \yes & \somewhat & \no  & \no  \\
Scideator             & \yes & \yes & \no        & \no  & \no  \\
Moose-Chem     & \yes & \no  & \no        & \no  & \no  \\
CodeScientist  & \yes & \yes & \yes       & \no  & \no  \\
AI researcher         & \yes & \yes & \yes       & \no  & \no  \\
%\rowcolor{gray!20}
\rowcolor[gray]{0.9}
\textbf{HARPA (ours)} & \yes & \yes & \yes       & \yes & \yes \\
\bottomrule
\end{tabular}
}
\caption{Comparison of ideation systems in terms of: 1) Are the ideas grounded in related work? 2) Can the ideator generate open-domain ideas? 3) Generates brief ideas or full proposal? 4) Does it consider feasibility w.r.t ASD agents? 5) Does it learn from prior experiments? (\yes: yes, \no: no, \somewhat: sometimes).\protect \footnotemark}
\label{tab:system_comparison}
\end{wraptable}

\footnotetext{Systems: GPT-5~\citep{openai2025gpt5},
Scideator~\citep{radensky2024scideator},
Moose-Chem~\citep{yang2024moose},
CodeScientist Ideator~\citep{jansen2025codescientist},
AI-Researcher~\citep{si2024can}
}

\paragraph{Automated ideation tools.} Computational frameworks such as Literature-Based Discovery (LBD)~\citep{swanson1986fish} illustrate how disconnected literatures can be bridged to reveal hidden hypotheses. More recent systems~\citep{radensky2024scideator,wang2023scimon} focus on producing super-brief, novelty-driven research ideas typically assessed with human judgments rather than execution. Systems such as Chain of Ideas~\citep{li2024chain} and HypER~\citep{vasu2025hyper} identify literature trends but generate ideas that are too high-level to be actionable, while Scideator~\citep{radensky2024scideator} generates diverse coarse-grained facets such as purpose, mechanism, or contribution, offering novelty but lacking operational clarity and require human refinement. IdeaSynth~\citep{pu2024ideasynth} transforms research ideas into proposals but demands substantial human-in-the-loop involvement, limiting scalability. Existing systems lack mechanisms to adapt their ideation in response to experimental feedback (Table~\ref{tab:system_comparison}). In this paper, we compare HARPA with the AI Researcher method~\citep{si2024can}, which was custom-built for open-domain proposal generation and has demonstrated state-of-the-art performance on this task.

\paragraph{Bridging ideation and execution.} Large-scale evaluations~\citep{si2024can, si2025ideation} show that while AI-generated ideas may be perceived as more novel than expert-authored ones, they are often less feasible experimentally. Other ideation frameworks, including MLR-Copilot~\citep{li2024mlrcopilotautonomousmachinelearning} and Agent Laboratory~\citep{schmidgall2025agentlaboratoryusingllm}, emphasize benchmark-guided or multi-agent settings but fall short of systematic experimental comparisons. Execution focused systems like \textsc{code scientist}~\citep{jansen2025codescientist} and AI-Scientist~\citep{lu2024ai} demonstrate end-to-end automated experimentation but assume hypotheses are already well-structured and feasible. HARPA complements these systems by generating structured, literature-grounded proposals. Unlike other ideators, HARPA integrates a reward model conditioned on ASD capabilities, making research hypotheses generation novel, grounded, and experimentally feasible. This makes it useful for human researchers positioning it as a building block toward the long-term vision of ``robot scientists"~\citep{doi:10.1126/science.1165620}.  Table \ref{tab:system_comparison} compares HARPA with representative systems in the literature over different ideation attributes.

%% file: chapters/method_harpa.tex
\section{HARPA: Hypothesis And Research Proposal Assistant}\label{sec:harpa}

HARPA's design is inspired by studies of how humans generate hypotheses (Section~\ref{sec:related-work}).
HARPA consists of two core components: a \emph{proposal generator} and a \emph{scorer}. The proposal generator begins with a user-given source paper and generates detailed, literature-grounded hypothesis-driven research proposals by treating hypotheses as structured research artifacts, enriched with a rationale (literature-based justification explaining how prior work motivates the preliminary hypothesis), related work, key variables, and operationalization plans. The scorer complements this process by ranking and filtering proposals with a learned reward model that predicts feasibility and testability without requiring full execution. These components together allow HARPA to produce hypothesis-driven proposals that are not only novel and grounded in prior work, but also prioritized for practical execution by ASD agents.

\begin{figure}[h]
    \centering
    \includegraphics[width=1.0\linewidth]{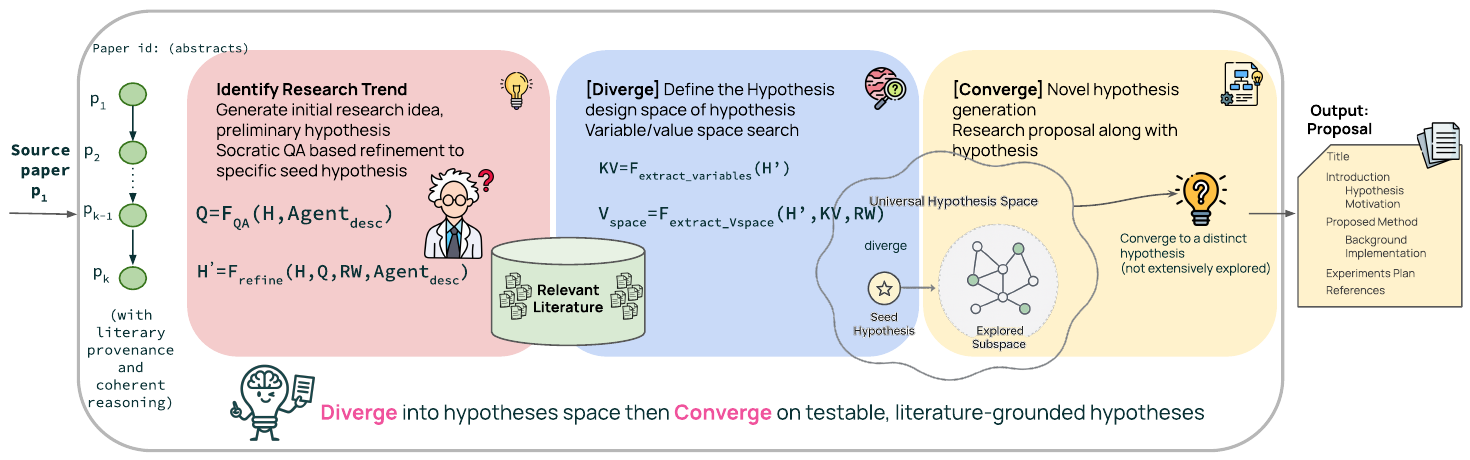}
    \caption{HARPA's Proposal Generator: Divergence and convergence to literature grounded novel proposals}
    \label{fig:harpa-overview-idea-gen}
\end{figure}

\subsection{HARPA's Proposal Generator
%: Divergence and convergence to literature grounded novel research proposals
}
 HARPA begins by constructing the scientific reasoning chain of papers given a source paper. The chain construction is based on~\citep{vasu2025hyper}, where each paper is connected to the previous paper based on its scientific dependency and the citing relation. This reasoning chain enables HARPA to identify a preliminary research gap and the motivation to come up with a research problem and hypothesis (Appendix L~\ref{app:preliminary_hyp}). However, the seed hypothesis generated at this stage is not very specific and is not optimized for novelty or the feasibility of the idea. To systematically develop a literature-grounded research proposal that is also novel and feasible, we follow the following steps:

 \paragraph{Refinement with Socratic Question Answering} Recently, Socratic questioning has been applied to language models as a self-guiding mechanism~\citep{chen2025socraticquestioninglearn}. We apply Socratic questioning to help the model think critically, uncover assumptions, and prompt a deeper understanding of the generic preliminary hypothesis. 
 Given a set of relevant literature snippets ($RW$) extracted from related works associated with this preliminary hypothesis, and the description of the underlying ASD agent ($Agent_{desc}$) which executes this hypothesis ($H$), we generate a set of at least 20 questions, $\mathcal{Q}$, that helps to navigate the specificity of this hypothesis (see Appendix L~\ref{app:SocraticQA_gen}). This is denoted as $\mathcal{Q} = \mathcal{F}_{QA}(H, Agent_{desc})$.
 Then, given this set of $\mathcal{Q}$ and the set of literature snippets $RW$ relevant to $H$ and $Agent_{desc}$, the language model can refine $H$ by answering these $\mathcal{Q}$. We denote this using $H^{'} = \mathcal{F}_{refine}(H, \mathcal{Q}, RW, Agent_{desc})$. The detailed instruction to refine and make the hypothesis more specific is given in Appendix L~\ref{app:refined_hypothesis}. Using  $H$ as the query, we systematically collect $RW$ using the snippet search over S2ORC corpus~\citep{singh-etal-2025-ai2}. Specifically, $H$ is progressively generalized $n$ times (see Appendix L~\ref{app:generalize_hypothesis}) and each version of $H$ is used as a query to collect $RW$.
  
 \paragraph{Defining the hypothesis design space using $H^{'}$} To understand the key concepts and variables around this hypothesis in hand, we first extract the set of key variables ($KV$) from it (see L\ref{app:variable_extraction}), denoted as $KV = \mathcal{F}_{extract\_var} (H^{'})$. However, the relevant literature might have already explored similar variables or different values of these variables (a brief illustration in Appendix L~\ref{app:variable_value_example}). We extract and define this key variable space as $\mathcal{V}_{space} = \mathcal{F}_{extract\_space} (H^{'}, KV, RW)$. In this process, we ensure that each of these key variables or values mentioned in the related literature is associated with $H^{'}$. To do this, the extraction process also extracts metadata such as the source paper title and the specific details and description relevant to $H^{'}$ about this variable (see Appendix L\ref{app:variable_space}). We also allow the model to add as extra variables a small set of standard evaluation metrics (``accuracy", ``precision") that were given as illustrative examples in the prompt. When these are added without direct literature evidence, they are explicitly marked as ``LLM-recommended". 
 %This way, we could trace back to the source paper or more details about the concepts that are described in the generated  proposal. 
 
\paragraph{Convergence to novel hypothesis} The research trend, initial idea, and hypothesis design space can be seen as HARPA's world model around the initial idea $H^{'}$. It encodes the key components of the idea and if or how they are being addressed in the relevant literature. Given this hypothesis space, $H^{'}$ is now converged into a distinct hypothesis, $H_{final}$, which has not been extensively studied in the given space. We denote this process as $H_{final} = \mathcal{F}_{generate} (H^{'}, \mathcal{V}_{space}, RW)$. Along with $H_{final}$, we also generate the detailed description of this hypothesis such as overview, detailed description of key variables, idea design including how the combination of the variables can be integrated or how the hypothesis can be implemented in a high level and some of the source papers (initial chain of papers and trend) from which this idea is evolved as related work (detailed instruction in  L\ref{app:final_hypothesis}, Appendix~\ref{appendix:prompts}). All LLM function calls in this pipeline were backed by \texttt{GPT-4o model}.

HARPA also specifies the operationalization of this idea, so that the underlying ASD agent or human researcher can have more details about its implementation plan. For this, we utilize the functionality---\texttt{idea to implementation plan}---of CodeScientist~\citep{jansen2025codescientist}. All this information together forms the final $\mathcal{HARPA}_{proposal}$ (example in Appendix~\ref{app:harpa_example}).

%% file: chapters/method_harpa_scorer.tex
%% proxy reward learner

\begin{figure}[t]
    \centering
    \includegraphics[width=1\linewidth]{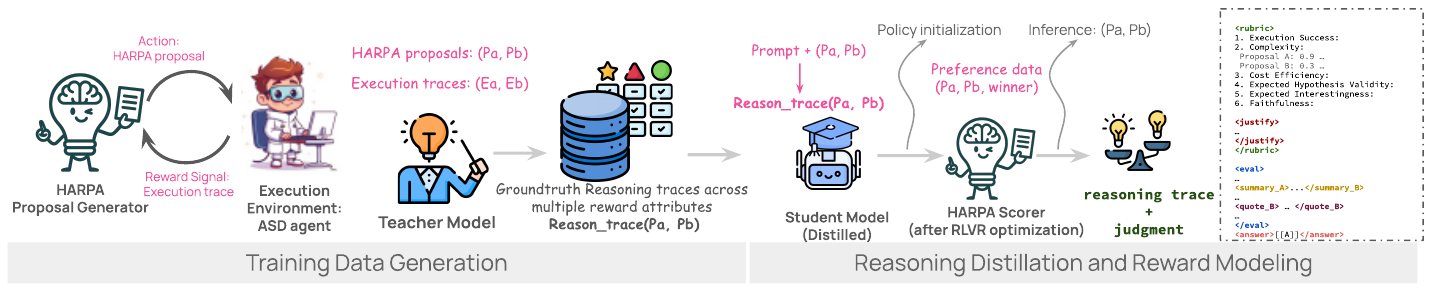}
    \caption{\textbf{HARPA Scorer:} \textit{1. Training Data Generation.} HARPA generates candidate proposals $(P_{a}, P_{b})$, which are executed in the ASD-agent environment to produce raw execution traces $(E_{a}, E_{b})$. A teacher LLM analyzes these traces and outputs a high-fidelity rubric-style reasoning trace with justification and answer $(Reason\_trace (P_{a}, P_{b}))$. \textit{2. Reasoning Distillation and Reward Modeling.} The student model is distilled from these reasoning traces, initialized as a policy, and fine-tuned via RLVR using preference labels to produce a rubric‑style reasoning trace and a preference label (e.g., ``Proposal A wins", an example trace in Appendix L~\ref{app:reasoning_trace}).
    %which can be used for downstream evaluation, interpretability, and research proposal refinement.
    }
    \label{fig:harpa-rm-r1-overview}
\end{figure}

\subsection{HARPA Scorer: Estimating testability of proposals}\label{subsec:scorer}
Generating and executing every candidate proposal, whether by human researchers or autonomous agents, is infeasible at scale. To address this challenge, we develop a \emph{learned reward model} that predicts the likely success of a research proposal without requiring full execution. 
Existing approaches either rely on direct execution (costly)~\citep{li2024mlrcopilotautonomousmachinelearning,lu2024ai} or on heuristic judgments by LLMs on feasibility~\citep{si2024can,Chen2025MLRBenchEA,yang2024moose,Baek2024ResearchAgentIR}, which are often unreliable~\citep{li2024llms} and lack grounding in prior experimental evidence~\citep{zhu2025ai}. Our goal is to provide a scalable and interpretable mechanism to filter and rank research proposals, prioritizing those that are both novel and feasible for the given ASD agent. See Figure~\ref{fig:harpa-rm-r1-overview} for the overview of the HARPA scorer.

\paragraph{Training Data Generation.} We collect preference data by executing HARPA-generated proposals using an off-the-shelf ASD agent, \textsc{CodeScientist}, that runs containerized Python experiments. Each execution ($E$) produces raw traces of the experiment setup, intermediate errors, and automatic assessments, and a final report. We convert the structured experiment summaries (e.g., Appendix L~\ref{app:cs_log_example}) generated by \textsc{CodeScientist} into categorical outcome labels using a meta-analysis scheme:
\[
\text{Label}(E) =
\begin{cases}
\begin{aligned}
&\text{Success} && \text{if } \texttt{faithfulness\_category} = \text{faithful}, \\
&\text{Failure} && \text{if } \texttt{faithfulness\_category} = \text{errors} \\
&&& \lor \ (\texttt{faithfulness\_category}=\text{inconclusive} \ \land \\ 
&&& \texttt{hypothesis\_category}=\text{inconclusive}), \\
&\text{Uncertain} && \text{otherwise.}
\end{aligned}
\end{cases}
\]
where \texttt{faithfulness\_category} indicates whether the experiment was executed faithfully without implementation errors, and \texttt{hypothesis\_category} captures whether the observed outcomes `support,' `reject,' or remain `inconclusive' w.r.t the original hypothesis. These labels are then used to construct pairwise preferences: for each pair $(P_a, P_b)$, a teacher LLM analyzed (see Appendix L~\ref{app:trace_generation}) the corresponding traces and generated a rubric-style reasoning trace, along with a preference judgment based on the observed outcome. This yields high-quality training data for the distillation, consisting of pairwise comparisons with interpretable justifications that reflect the empirical feasibility.

\paragraph{Reasoning Distillation and Reward Modeling.} We train the HARPA scorer in two stages following the RM-R1 framework~\citep{chen2025rm}. First, we distill the teacher's rubric-style reasoning traces into the student model. This facilitates the student with the ability to generate interpretable justifications aligned with teacher rubrics. Next, we train the distilled model with preference-based optimization using the RLVR strategy~\citep{chen2025rm}, aligning its scoring with empirically verifiable outcomes (`success,' `failure') from \textsc{CodeScientist} executions. The model outputs both (i) a comparative label (e.g., ``Proposal A wins''
) and (ii) a rubric-style reasoning trace explaining the decision. This dual output allows the model to function not only as a black-box scorer but also as an explainer, providing transparent, human-readable justifications that can be used to refine research proposals. An example reasoning trace is provided in Appendix L~\ref{app:reasoning_trace}, showing how the model assigns higher feasibility to one proposal using execution-derived factors, such as execution success, complexity (based on reflection), and cost efficiency parsed from the structured experiment summary of \textsc{CodeScientist}.

\paragraph{Conditioning on ASD capabilities.} To ensure judgements are adaptable to the targeted execution environment, the reward model is conditioned on an explicit ASD agent (see Appendix L~\ref{app:trace_generation}), specifying constraints such as compute budget, permissible evaluation protocol, dataset access, and whether human involvement is allowed. During both training and inference, the agent profile is concatenated with the proposals and execution metadata. In our case, conditioning reflects the limits of \textsc{CodeScientist}, but the same mechanism applies to other agents. For instance, proposals requiring human studies or private datasets are down-ranked for \textsc{CodeScientist} but could go higher for a more capable agent. This makes HARPA's scorer adaptive, producing feasibility-aware rankings that generalize across different discovery settings.

%% file: chapters/experiments.tex
\section{Experiments}\label{sec:experiments}
We evaluate HARPA along two complementary axes: (1) a human-centric expert study to evaluate whether generated proposals are appealing to human researchers, and (2) an ASD-centric execution study, which measures the operational testability of proposals through the reward modeling. 

\subsection{Baselines}\label{sec:baseline}

We compare HARPA against different baselines depending on the evaluation axis. 
\textbf{Human-centric Evaluation:} We compare HARPA proposal generator against AI-Researcher~\citep{si2024can}, a strong baseline for literature-grounded ideation. 
We standardized section headings to match proposal formats across systems. For references, we included the papers AI-Researcher internally retrieved, whereas HARPA had literature identified during its multi-stage pipeline. To ensure comparability, we generated topics from each source paper's abstract (since AI-Researcher expects a topic rather than a source paper). Apart from this topic generation step, all other settings followed the original AI-Researcher implementation. 
\textbf{Agent-centric Evaluation:} For the HARPA scorer, we compare the two variants: (i) an untrained LLM scorer applied directly to a pair of proposals, and (ii) the HARPA scorer, our distilled and RLVR-trained reward model. This setup allows us to isolate the benefit of training the scorer while keeping the proposal generator fixed. We use \texttt{Qwen-7B-Instruct} as the backbone, with the non-finetuned model as the LLM scorer baseline and the trained version as HARPA scorer.

\subsection{Human-centric Evaluation Setup}\label{sec:human-exp}

\paragraph{Participants:} We recruited $12$ experts who have experience in writing and reviewing scientific articles in their domain of interest via \emph{Upwork.com}. See Appendix~\ref{app:human_eval_details} for detailed backgrounds and screening criteria. 
\textbf{Dataset:} Our evaluation corpus was constructed dynamically by the experts themselves. Each expert selected source papers ($\geq 20$ citations, published before 2025) in their domain of expertise. This design ensured informed and fair evaluation in a familiar context. For each source paper, we generated two proposals from HARPA and two from the baseline, and each expert evaluated proposals from at most two source papers of their choice. This process resulted in $40$ proposals per system overall. Proposals were uniformly formatted with identical section headings --- \texttt{title}, \texttt{introduction}, \texttt{proposed method}, \texttt{experiments plan}, and \texttt{references}, and covered diverse topics (e.g., NLP, RAG, RL, Optimization). (Corpus statistics in Appendix~\ref{app:topic_info})

\paragraph{Evaluation Rubric:} We adapted our evaluation rubric from the idea review form of~\cite{si2024can} for evaluating research proposals. Experts rated each proposal on a 10-point Likert scale for \textit{Familiarity, Novelty, Feasibility, Expected effectiveness, Excitement, Overall, and Confidence}, providing brief textual justifications (full rubric in Appendix~\ref{app:review_form}). In addition to the original rubric, we introduced four dimensions relevant to hypothesis-driven proposals and their operationalization: \textit{Literature Grounding, Motivation from Literature, Coherence of Idea Composition, and Specificity of Proposed Method}. In total, the rubric covered $11$ dimensions, with full wording provided in Appendix~\ref{app:review_form}. \textbf{Protocol:} Proposals were presented to each expert in randomized order, with system identities hidden. 
The same expert who provided the source paper independently assessed and rated all four proposals (including baseline and HARPA) to ensure fair comparison on the same topic. Some experts reviewed proposals for more than one source paper
\footnote{Since source papers were selected individually, proposals were unique to each expert and not cross-reviewed.}. Data collection was carried out using the Label Studio platform and experts were compensated at a rate of 35USD/hr.

\subsection{Agent-centric Evaluation Setup}\label{sec:agent-exp}
Here we evaluate proposals by executing them with the \textsc{CodeScientist} providing data.

%for training and testing HARPA scorer.
\paragraph{Data Curation.} We sampled $275$ highly cited ACL papers as source papers and generated up to five HARPA proposals per paper ($1,222$ total). Each proposal was executed \emph{five} times each in \textsc{CodeScientist} to avoid the stochasticity in LLM-based code generation. From each of the five runs, we selected the execution trace that most truly representing the research proposal and considered that for further analysis. Outcomes were labeled as \textsc{Success} ($29.38\%$), \textsc{Failure} ($51.55\%$), or \textsc{Uncertain} ($19.07\%$) according to the categorical outcome labels described before.

\paragraph{Preference Construction and Training:} From these labeled executions, we constructed $3954$ preference pairs on shared source paper topic (see Appendix L~\ref{app:source_paper_topic}). Each pair with execution metadata was used to generate a rubric-style reasoning trace (including preference judgments) by an oracle model\footnote{\texttt{claude-sonnet-4} was used as an Oracle model.}, which achieved $87.48\%$ accuracy. We filtered the pairs with correct judgements ($3459$) and their reasoning traces as ground truth for further experiments. 
\textbf{Distillation and RLVR:} We split the proposals into training ($2595$), validation ($452$), and test ($412$) subsets. Following the RM-R1 framework~\citet{chen2025rm}, we first distilled a student model to generate interpretable rubric-style reasoning aligned with teacher rubrics. We further applied RLVR training on preference pairs (using an additional $226{,}170$ success-failure pairs irrespective of shared topic). Finally, we evaluated the distilled reward model on a held-out set of success-failure pairs ($186$), using accuracy and qualitative analysis of reasoning traces. This two-stage process yields the \textit{HARPA Scorer} that is both interpretable and adaptive to ASD execution (more implementation details in Appendix~\ref{app:implementation}).

\begin{figure}[tbh]
    \centering
    \includegraphics[width=0.6\linewidth]{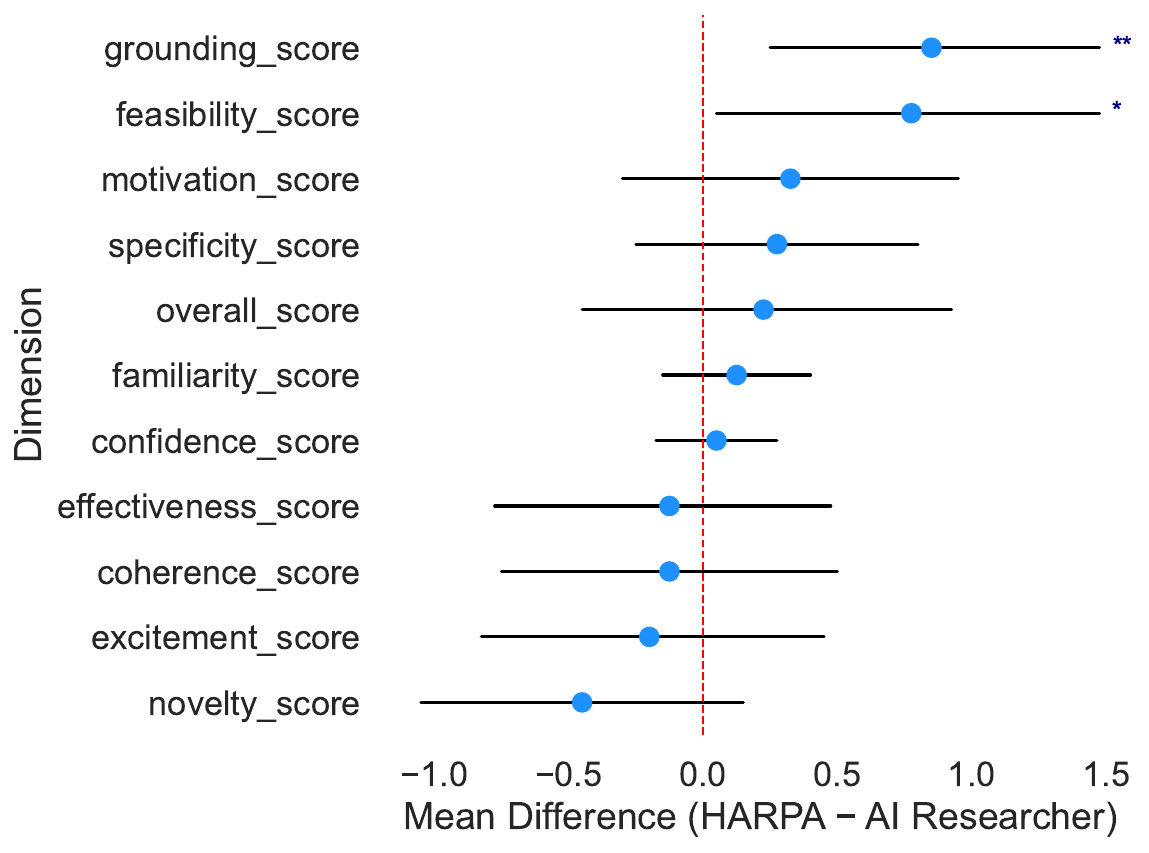}
    \caption{Mean difference between HARPA's proposal generator and AI-Researcher across nine evaluation dimensions. Also reporting the familiarity and confidence score differences. Points show average differences, horizontal bars indicate 95\% bootstrap confidence intervals ($10k$ resamples). Stars indicate significant difference computed using the nonparametric bootstrap test (* $p<0.05$, **$p<0.01$)}
     \label{fig:mean_diff_ci}
\end{figure}

A full end-to-end evaluation would be ideal, but it is too expensive and requires impractical expert annotation of random proposals. We instead combine expert review on a small set of relevant samples with large-scale testability on diverse ACL papers, leaving full evaluation to future work.

%% file: chapters/results.tex
\section{Main results}\label{sec:results}

\subsection{Human-centric Results}

Figure~\ref{fig:mean_diff_ci} summarizes the expert evaluations of HARPA's proposal generator against the baseline across $11$ dimensions. Nine dimensions define the research proposal quality (i.e., novelty, feasibility, expected effectiveness, excitement, grounding, specificity, coherence, motivation, and overall quality), while two meta-dimensions capture the user's familiarity with the proposal topic and their confidence in the judgment (complete proposal evaluation form in Appendix~\ref{app:review_form}). HARPA shows statistically significant gains in feasibility (+0.78, $p<0.05$, bootstrap) and grounding (+0.85, $p<0.01$, bootstrap). For specificity, motivation, and overall scores, HARPA shows a positive trend, although it does not rise to the level of statistical significance. For other metrics, HARPA performs comparably to the baseline (Appendix Table~\ref{tab:expert_scores_full}), showing that improvements in feasibility and grounding without sacrificing clarity or novelty. Novelty scores for HARPA averaged $5.98 \pm 1.33$ compared to $6.43 \pm 1.32$ for the baseline, with both systems rarely falling below the midpoint of the 10-point scale. This indicates that HARPA produces ideas perceived as incrementally novel. These findings align with our design goal that grounding research proposals in literature and refining hypotheses through a human-like workflow leads to more operational, testable research proposals. (Detailed  rating distributions in Appendix~\ref{app:human_eval_details}.)

%only keep one figure?
\begin{figure}[tbh]
  \centering
  \begin{subfigure}[t]{0.48\linewidth}
    \centering
    \includegraphics[width=\linewidth]{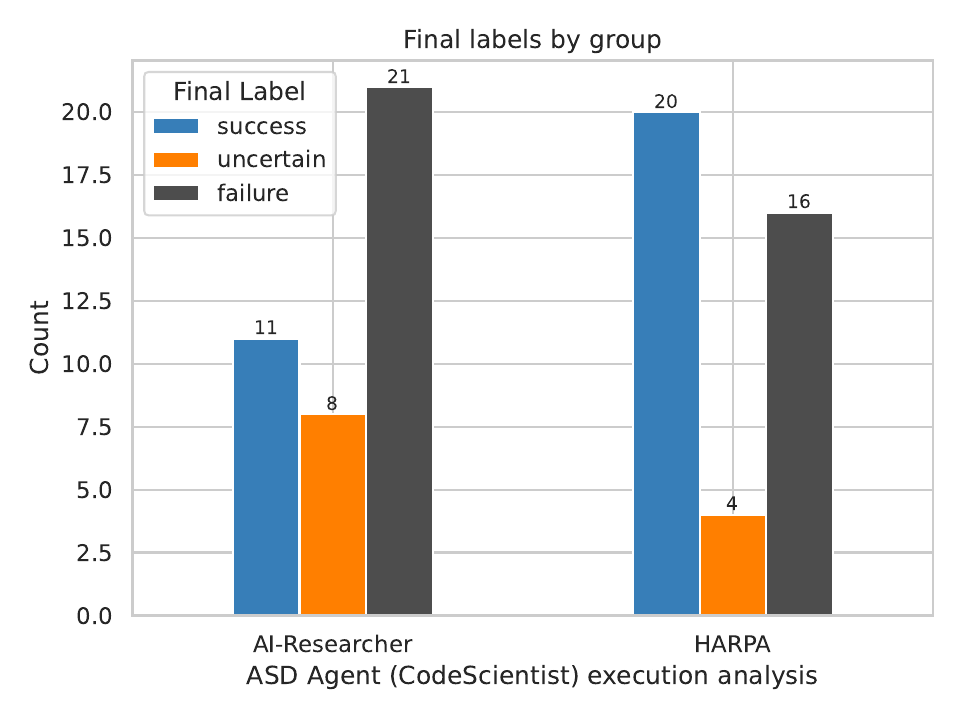}
    \caption{Execution outcomes (counts) from \textsc{CodeScientist} runs, 
    labeled as success, uncertain, or failure.}
    \label{fig:ASD_counts}
  \end{subfigure}\hfill
  \begin{subfigure}[t]{0.48\linewidth}
    \centering
    \includegraphics[width=\linewidth]{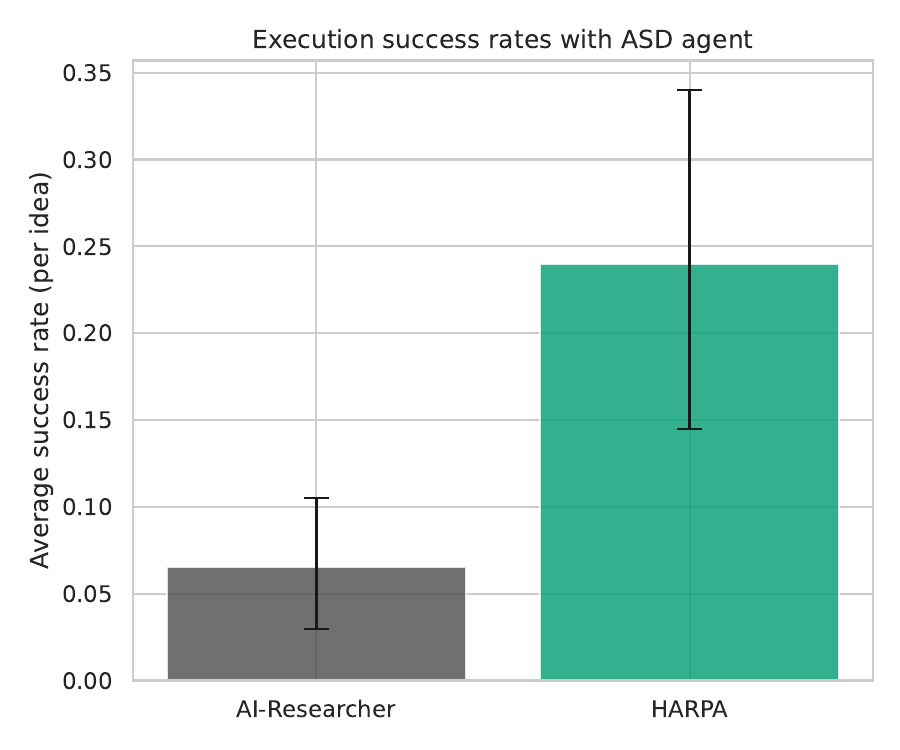}
    \caption{Average per-idea success rates (95\% bootstrap CIs) for HARPA 
    vs.\ baseline AI-Researcher proposals.}
    \label{fig:ASD_successCI}
  \end{subfigure}
  \caption{Execution results from \textsc{CodeScientist}. 
  Left: outcome distribution across groups. 
  Right: paired comparison of mean success rates showing HARPA significantly outperforms the baseline AI-Researcher.}
  \label{fig:ASD_feasibility}
\end{figure}

\paragraph{Execution Success rates:} We evaluated whether HARPA proposals more often succeed when executed by a typical ASD agent (here, \textsc{CodeScientist}). For each user-evaluated proposal, we executed five independent runs labeled outcomes using the meta-analysis labeling scheme (Section~\ref{subsec:scorer}). Figure~\ref{fig:ASD_counts} shows the raw distribution of execution outcomes across groups. HARPA produced a higher number of successful executions (20 vs.\ 11 for the baseline) and fewer outright failures than the baseline AI-researcher system.  We further aggregated results into per-idea success rate, defined as the proportion of faithful executions per idea. To ensure fair comparison, we paired HARPA and baseline proposals generated from the same source paper and computed within-source paper differences in success rates. Figure~\ref{fig:ASD_successCI} summarizes per-idea success rates where HARPA achieved a higher mean success rate than the baseline (0.24 vs.\ 0.065), and the bootstrap test ($p<0.001$) confirmed the difference was statistically significant. These results together demonstrate that HARPA proposals succeed more often in automated executions, consistent with expert ratings of higher feasibility.

\begin{table}[t]
\centering
\small
\renewcommand{\arraystretch}{1.2}
\setlength{\tabcolsep}{4pt} % tighter spacing
\begin{tabular}{lccccccc}
\toprule
\multicolumn{1}{c}{\textbf{System}} & 
\multicolumn{6}{c}{\textbf{Pairwise Consistency}} & 
\multicolumn{1}{c}{\textbf{Accuracy}} \\
\cmidrule(lr){2-7} 
& \makecell{Execution \\ Success} & 
\makecell{Complexity} & 
\makecell{Cost \\ Efficiency} & 
\makecell{Hypothesis \\ Validity} & 
\makecell{Interesting- \\ ness} & 
\makecell{Faithfulness} & 
\makecell{(win)} \\
\midrule
Baseline & -- & -- & -- & -- & -- & -- & 0.52 \\
HARPA    & 0.80 & 0.67 & 0.73 & 0.69 & 0.55 & 0.79 & \textbf{0.81} \\
\bottomrule
\end{tabular}
\caption{Pairwise consistency of \textbf{HARPA-Scorer} with oracle judgments across rubrics. Consistency is the fraction of proposal pairs where the scorer and oracle agree. Baseline lacks rubric-level judgments (--). Accuracy comparing baseline and \textbf{HARPA-Scorer} on the test data (N=186 success-failure pairs).}
\label{tab:harpa-rm}
\end{table}

\begin{table}[t]
\centering
\small
\begin{tabular}{lccccccc}
\toprule
System & BLEU & BLEU BP & ROUGE-1 & ROUGE-2 & ROUGE-L & ROUGE-Lsum & Len-ratio \\ % & Gen-len & Ref-len \\
\midrule
Baseline  &  0.08 &  0.79     &  0.43     &  0.13     &  0.18     &  0.17      &  0.81        \\ %&  125,739 &  155,088 \\
HARPA &  0.22 &  1.00 &  0.55 &  0.22 &  0.26 &  0.26  &  1.12   \\ % &  174,365 &  155,088 \\
\bottomrule
\end{tabular}
\caption{Overlap-based evaluation of HARPA reasoning traces w.r.t reference traces.}
\label{tab:harpa-rationale-scores}
\end{table}

\paragraph{Qualitative Examples:} To illustrate these quantitative trends, an expert rated a HARPA proposal as highly feasible and grounded: ``\textit{Using the softmax trick allows backpropagation/gradient estimation, it is a well known trick and the implementation is not so complicated...}" (feasibility = 7, grounding = 9). The expert highlighted that the ideas was concrete and testable, with direct support from the prior literature. By contrast, a baseline proposal as ``\textit{The proposed method looks feasible. The problem is that it lacks details. Everything related to the method is summarized in 2--3 lines in the `Proposed methods' without any mathematical language. ...}" was judged infeasible (feasibility = 4) and poorly grounded (grounding = 2). Although the expert noted that it was an exciting impact, they emphasized that the lack of detail and irrelevant literature made the proposal impossible to operationalize. Examples of full proposals and expert assessments are provided in Appendix~\ref{app:expert_examples}

\noindent
\textbf{\faSearch~Takeaway:} \textit{In summary, HARPA bridges the gap between ideation and execution: it generates literature-grounded, feasible, and testable research proposals that succeed \textbf{nearly twice as often in ASD execution (20 vs.\ 11, $\approx$2$\times$)}, while also outperforming prior systems in expert evaluations.}

\subsection{Agent-centric Results}

We next evaluated HARPA-scorer against a baseline untrained LLM scorer, a \texttt{Qwen-7B-instruct} (section~\ref{sec:baseline}). HARPA-scorer improves accuracy with a +0.28 absolute gain (a 53\% relative improvement), with more balanced performance across classes. %20\% cost saving comapred to random baseline 
In addition, HARPA-scorer produces rubric-aligned reasoning traces with explicit scoring on feasibility, cost efficiency, and complexity, like the teacher model. In contrast, the baseline model produced unstructured free text (e.g., in Appendix L~\ref{app:baseline_reasoning_trace}) that lacks actionable justifications and $4.84\%$ unknown predictions. This alignment with oracle-style reasoning makes HARPA-scorer's judgement easier to interpret and more reliable for refining the research proposals.

Beyond accuracy, we further assessed \textit{pairwise consistency}, whether the scorer agrees with the Oracle on which two proposals are preferred for each rubric dimension. HARPA-scorer achieves strong alignment on testability-oriented rubrics, with $0.80$ consistency on \textit{Execution Success} and $0.70$ on \textit{faithfulness}, and moderate alignment on \textit{Complexity, Cost Efficiency, and Hypothesis Validity}, while alignment drops to $0.55$ for the more subjective \textit{Interestingness} dimension. These results indicate that the scorer is capturing reliably testability-related signals while remaining less consistent on subjective criteria. Finally, we compare HARPA-generated rationales with baseline ones. HARPA significantly ($p<0.01$, paired t-test) outperforms the baseline across all overlap metrics with reference rationales. We see major improvements in BLEU scores (+0.14, a 166\% increase) and strong gains in ROUGE-1 (+0.12, +27\%), ROUGE-2 (+0.10, +77\%), and ROUGE-L/Lsum (+0.09, +49\%). The particularly strong improvements in higher-order n-grams—like ROUGE-2 and BLEU's 3-4-gram scores—suggest that HARPA is not just matching individual words better, but is actually producing more coherent text with better content flow and sequencing.

\noindent
\textbf{\faSearch~Takeaway:} \textit{In summary, HARPA's scorer delivers \textbf{+0.28 absolute ($\sim$53\% relative) higher accuracy} than an untrained LLM scorer, while providing interpretable rubric-style judgments that enable reliable, execution-informed filtering of research proposals.}

%% file: chapters/conclusion.tex
\section{Conclusion and Future Work}
We presented HARPA, a literature-grounded, testability-driven framework for the open-ended task of hypothesis generation. HARPA systematically extracts research trends, explores existing hypothesis spaces, and converges on testable hypothesis-driven proposals. We introduced an interpretable reward-trained scorer that adapts feasibility judgments to ASD agent capabilities, enabling HARPA to prioritize hypotheses that are executable. Our evaluations show significant improvements in feasibility and grounding, with HARPA's proposals also succeeding more often in automated execution. As the scorer serves as a proxy for resource-intensive experimentation, HARPA enables execution-derived feedback into future proposal generation, selectively refining hypotheses in line with ASD agent capabilities.
To our knowledge, HARPA is the first ideation framework to learn directly from execution outcomes, enabling feasibility-aware hypotheses generation, and points to further possible improvements using even richer training data and execution environments. Together, these contributions represent a step toward more capable hypothesis generation tools and help advance the rapidly growing field of AI-driven scientific discovery.

%% file: appendix/human_eval.tex
\subsection{Human Evaluation Details}\label{app:human_eval_details}

We recruited 12 experts with demonstrated research experience, spanning diverse academic and research backgrounds. The expert pool included 6 PhD students in Computer Science or related fields, 3 faculty members/academic researchers, and 2 postdoctoral researchers. Their expertise covered a broad range of topics in ML and NLP: bias and fairness in language models (4), multilingual and low-resource NLP (3), factuality and hallucination detection (3), code generation and programming with LLMs (2), uncertainty estimation and interpretability (2), and mathematical reasoning/structured predictions (2). Additional specialized domains included recommender systems and IR, mathematical modeling, deep reinforcement learning, and AI safety/robustness. Note that the counts are not mutually exclusive, as evaluators could select multiple primary research areas.

In terms of research experience, half of the participants (6/12) reported 3-5 years of active work in their field, three reported 6-10 years, and three reported 10+ years. As part of the screening, each expert shared their Google Scholar profile (or equivalent evidence of publications). The citation count of experts' scholarly work ranged from 7 to 1256 (median =147, mean=297.3). This distribution shows that our evaluation pool included both early-career researchers and more senior researchers with substantial publication records. 

\subsubsection{Source paper domains and topic distribution}\label{app:topic_info}

\begin{figure*}[h]
    \centering
    \begin{subfigure}{0.32\textwidth}
        \centering
        \includegraphics[width=\linewidth]{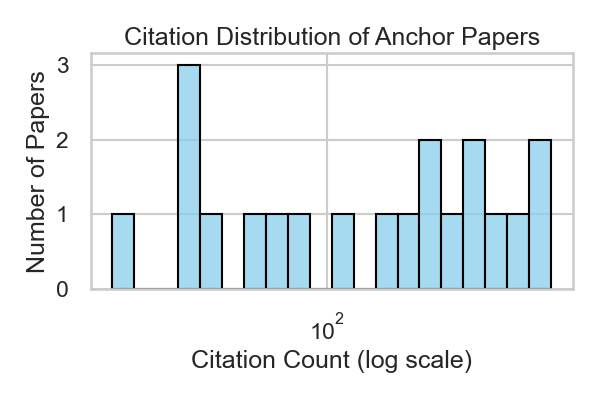}
        \caption{Citation distribution}
        \label{fig:anchor_citations}
    \end{subfigure}
    \hfill
    \begin{subfigure}{0.32\textwidth}
        \centering
        \includegraphics[width=\linewidth]{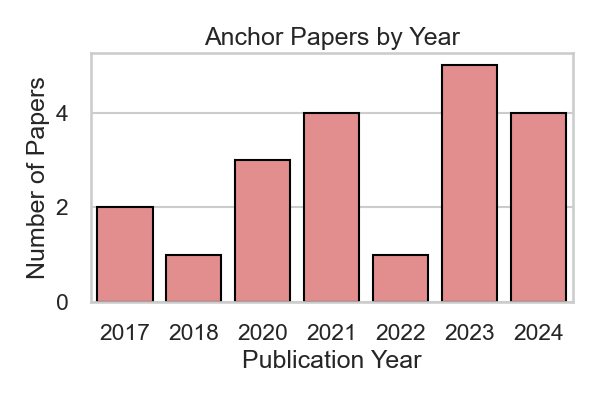}
        \caption{Publication years}
        \label{fig:anchor_years}
    \end{subfigure}
    \hfill
    \begin{subfigure}{0.32\textwidth}
        \centering
        \includegraphics[width=\linewidth]{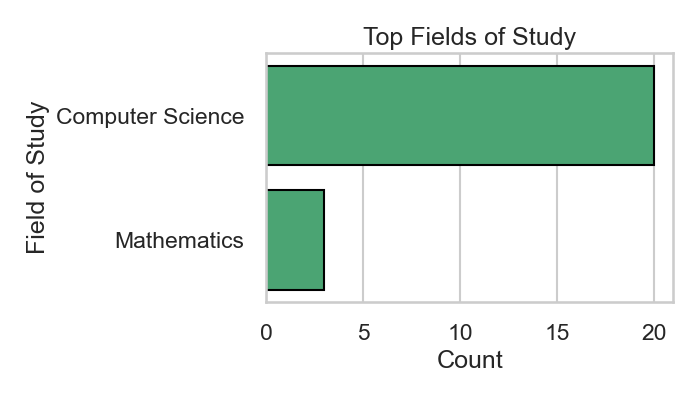}
        \caption{Top fields of study}
        \label{fig:anchor_fields}
    \end{subfigure}
    \caption{Aggregate statistics of source papers selected by experts.}
    \label{fig:anchor_summary}
\end{figure*}

Table~\ref{tab:anchor_topics} summarizes the distribution of source papers across shared topics, obtained by classifying abstracts into broad topics using the same approach from Listing~\ref{app:source_paper_topic}. The topics span from graph neural networks to health-care applications, optimization to fairness. Figure~\ref{fig:anchor_summary} and Table~\ref{tab:anchor_venues} summarize the source papers selected by experts. These papers span recent years (2018-2023), show moderate citation counts, and cover diverse venues. 

\begin{table}
\caption{Distribution of source papers across shared topics.}
\label{tab:anchor_topics}
\begin{tabular}{lr}
\toprule
 & Count \\
shared_topic &  \\
\midrule
Graph Neural Networks and Graph Learning & 3 \\
Generative Models for Images & 2 \\
Recommender Systems with LLMs & 2 \\
Medical and Healthcare Applications & 2 \\
Differentiable Optimization & 2 \\
Bias and Fairness in NLP & 2 \\
Hallucination and Factuality in LLMs & 2 \\
Reinforcement Learning for Scheduling & 2 \\
Retrieval-Augmented Generation and Information Refinement & 1 \\
Continual Learning and Knowledge Distillation & 1 \\
Fake News Detection & 1 \\
\bottomrule
\end{tabular}
\end{table}

\begin{table}[h]
\centering
\small
\begin{tabular}{lc}
\toprule
\textbf{Venue} & \textbf{Count} \\
\midrule
arXiv.org & 3 \\
Neural Information Processing Systems & 2 \\
Annual Meeting of the Association for Computational Linguistics & 2 \\
Computer Vision and Pattern Recognition & 2 \\
North American Chapter of the Association for Computational Linguistics & 2 \\
International Conference on Computer Graphics and Interactive Techniques & 1 \\
IEEE Access & 1 \\
Knowledge Discovery and Data Mining & 1 \\
ACM Transactions on Intelligent Systems and Technology & 1 \\
ACM Conference on Health, Inference, and Learning & 1 \\
\bottomrule
\end{tabular}
\caption{Venues of source papers selected by experts.}
\label{tab:anchor_venues}
\end{table}

\begin{figure}[t]
    \centering
    \includegraphics[width=\linewidth]{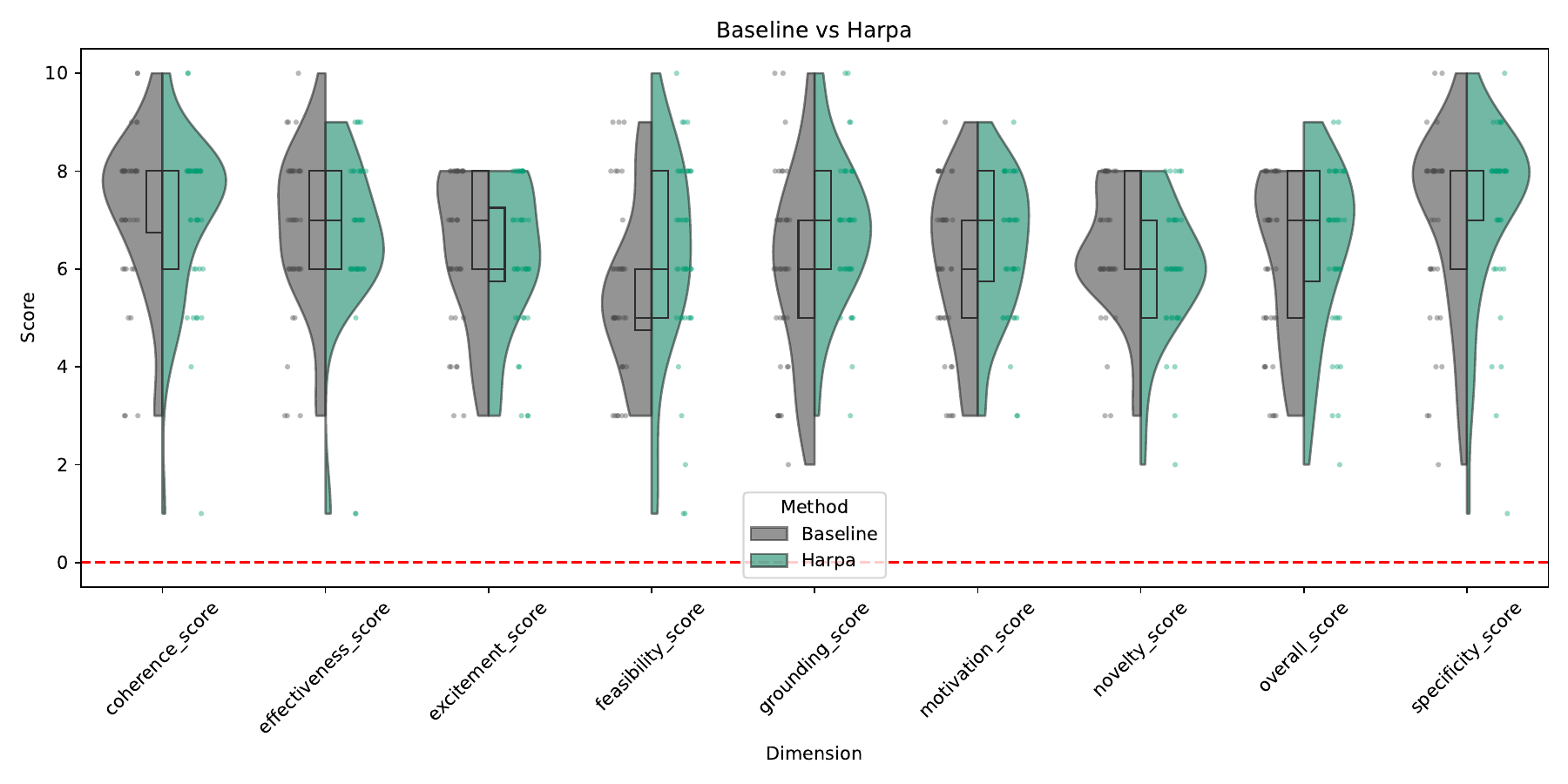}
    \caption{Distribution of expert ratings across nine dimensions for HARPA vs baseline. Shown for completeness (complementary to Fig.~\ref{fig:mean_diff_ci}).}
    \label{fig:appendix_violin}
\end{figure}

\newpage
\section{Proposal Review Form}
\label{app:review_form}

We use the following proposal assessment form to elicit reviews from all the experts. Our assessment questions largely follow the expert evaluation protocol introduced by~\cite{si2024can} for proposal assessment, but we extended it with several additional questions tailored to hypothesis-driven research proposals. In particular, we added dimensions for \textit{motivation}, \textit{specificity}, \textit{coherence}, and \textit{literature grounding}, as these aspects are critical for ensuring that proposals are both operational and directly testable. To ensure consistency, each question was accompanied by a detailed description of the scale points as well as hints on what evidence to consider (e.g., explicit references, prior knowledge). 

The full questionnaire (including all Likert-scale anchors and instructions shown to experts) is reproduced below. 

\textbf{\textcolor{purple}{1. Familiarity}}: Before reviewing the idea, please indicate how familiar you are with the given topic on a scale of 1 - 5 (this is just for us to understand potential confounders). 

\begin{enumerate}
    \item You have never read about this topic before
    \item You have read at least one paper on this topic
    \item You have read multiple papers on this topic but have not published any paper on it
    \item You have co-authored at least one paper on this topic
    \item You have co-authored multiple papers on this topic or have published at least one first-author paper on this topic
\end{enumerate}

\textbf{\textcolor{purple}{2. Novelty Score}}: Whether the idea is creative and different from existing works on the topic, and brings fresh insights. You are encouraged to search for related works online. You should consider all papers that appeared online prior to July 2024 as existing work when judging the novelty.

\begin{enumerate}
    \item Not novel at all - there are many existing ideas that are the same
    \item 
    \item Mostly not novel - you can find very similar ideas
    \item 
    \item Somewhat novel - there are differences from existing ideas but not enough to turn into a new paper
    \item Reasonably novel - there are some notable differences from existing ideas and probably enough to turn into a new paper
    \item 
    \item Clearly novel - major differences from all existing ideas
    \item 
    \item Very novel - very different from all existing ideas in a very interesting and clever way
\end{enumerate}

\textbf{\textcolor{purple}{Novelty Rationale}}: Short justification for your score. If you give a low score, you should specify similar related works. (Your rationale should be at least 2-3 sentences.)
\emph{Hint: If the idea is not novel, point out what is already similar in prior work (e.g., method, task, or setting), and briefly mention any minor differences if they exist. If the idea is novel, explain what is new—such as a novel method, a new task, or applying an existing idea to a new domain.}

\textbf{\textcolor{purple}{3. Feasibility Score}}: How feasible it is to implement and execute this idea as a research project? Specifically, how feasible the idea is for a typical CS PhD student to execute within 1-2 months of time. You can assume that we have abundant OpenAI / Anthropic API access, but limited GPU compute.

\begin{enumerate}
    \item Impossible: the idea doesn't make sense or the proposed experiments are flawed and cannot be implemented
    \item 
    \item Very challenging: there are flaws in the proposed method or experiments, or the experiments require compute/human resources beyond any academic lab
    \item 
    \item Moderately feasible: It can probably be executed within the given time frame but would require careful planning, efficient use of APIs or some advanced computational strategies to overcome the limited GPU resources, and would require some modifications to the original proposal to make it work
    \item Feasible: Can be executed within the given constraints with some reasonable planning
    \item 
    \item Highly Feasible: Straightforward to implement the idea and run all the experiments
    \item 
    \item Easy: The whole proposed project can be quickly executed within a few days without requiring advanced technical skills
\end{enumerate}

\textbf{\textcolor{purple}{Feasibility Rationale}}: Short justification for your score. If you give a low score, you should specify what parts are difficult to execute and why. (Your rationale should be at least 2-3 sentences.)

\textbf{\textcolor{purple}{4. Expected Effectiveness Score}}: How likely the proposed idea is going to work well (e.g., better than existing baselines).

\begin{enumerate}
    \item Extremely Unlikely: The idea has major flaws and definitely won't work well
    \item 
    \item Low Effectiveness: The idea might work in some special scenarios but you don't expect it to work in general
    \item 
    \item Somewhat ineffective: There might be some chance that the proposed idea can work better than existing baselines but the improvement will be marginal or inconsistent
    \item Somewhat effective: There is a decent chance that the proposed idea can beat existing baselines by moderate margins on a few benchmarks
    \item 
    \item Probably Effective: The idea should offer some significant improvement over current methods on the relevant benchmarks
    \item 
    \item Definitely Effective: You are very confident that the proposed idea will outperform existing methods by significant margins on many benchmarks
\end{enumerate}

\textbf{\textcolor{purple}{Expected Effectiveness Rationale}}: Short justification for your score. (Your rationale should be at least 2-3 sentences.) \textit{Hint: You must consider how the novelty of the idea relates to its excitement or impact — if the idea is not novel (e.g., already done before), it should generally not be rated as very exciting.}

\textbf{\textcolor{purple}{5. Excitement Score}}: How exciting and impactful this idea would be if executed as a full project. Would the idea change the field and be very influential.

\begin{enumerate}
    \item Poor: You cannot identify the contributions of this idea, or it's not interesting at all and you would fight to have it rejected at any major AI conference
    \item 
    \item Mediocre: this idea makes marginal contributions and is very incremental
    \item 
    \item Leaning negative: it has interesting bits but overall not exciting enough
    \item Learning positive: exciting enough to be accepted at a major AI conference, but still has some weaknesses or somewhat incremental
    \item 
    \item Exciting: would deepen the community's understanding or make major progress in this research direction
    \item 
    \item Transformative: would change the research field profoundly and worth a best paper award at major AI conferences
\end{enumerate}

\textbf{\textcolor{purple}{Excitement Rationale}}: Short justification for your score. (Your rationale should be at least 2-3 sentences.) \textit{Hint: You must consider how the novelty of the idea relates to its excitement or impact — if the idea is not novel (e.g., already done before), it should generally not be rated as very exciting.}

\textbf{\textcolor{purple}{6. Literature Grounding}}: To what extent are the key components (e.g., model choice, tasks, evaluation strategies) grounded in existing scientific literature? You may also consider whether ideas reflect well-established domain knowledge or listed references.

\begin{enumerate}
    \item Not at all grounded: Mostly speculative or hallucinated; no support from literature or well-established concepts
    \item 
    \item Weak grounding: A few connections to existing work, but most claims lack clear support from the listed references or alignment with well-known concepts
    \item 
    \item Partially grounded: About half the components are linked to literature or reflect widely accepted ideas in the field
    \item Strong grounding: Most core elements are supported by the listed references or well-established concepts, with only minor gaps
    \item 
    \item Very strong grounding: The vast majority of components are supported by listed references or widely accepted domain knowledge, though one or two key claims still lack clear support
    \item 
    \item Fully grounded: Every major concept and step is well-supported by listed references or clearly based on well-established domain knowledge; no unsupported claims remain.
\end{enumerate}

\textbf{\textcolor{purple}{Explanation}}: You should also provide a rationale for your score. (Your rationale should be at least 2-3 sentences.) 
\textit{Hint: If a claim is grounded in well-known concepts but not supported by the listed references, explain why it is reasonable based on your domain knowledge. Indicate whether your assessment relies on (a) the proposal’s reference list, (b) external sources you know, or (c) generally accepted field knowledge.}

\textbf{\textcolor{purple}{7. Motivation from Literature}}: Is the problem statement/overall idea clearly defined and motivated by a specific, well-scoped research gap, or limitation identified in the widely recognized field knowledge?

\begin{enumerate}
    \item No clear motivation: idea feels arbitrary or disconnected
    \item 
    \item Weakly motivated: mentions general themes but lacks a compelling rationale
    \item 
    \item Somewhat motivated: a recognizable problem is present, but vague
    \item Well motivated: builds on a clear and relevant research direction
    \item 
    \item Strongly motivated: clearly addresses a known issue or opportunity from existing work or widely acknowledged field challenges
    \item 
    \item Exceptionally motivated: makes a compelling case for a timely and important problem grounded in the reference list or broadly recognized research needs
\end{enumerate}

\textbf{\textcolor{purple}{Explanation}}: You should also provide a rationale for your score. (Your rationale should be at least 2-3 sentences.) Also specify which part of the idea was most clearly linked to a literature-based motivation.

\textbf{\textcolor{purple}{8. Coherence of Idea Composition}}: Are the combined components (problem, methods, tasks, and metrics) logically integrated and literature-informed?

\begin{enumerate}
    \item Incoherent: parts don’t fit together; lacks logical or conceptual connection
    \item 
    \item Loosely connected: some rationale exists, but combination feels forced
    \item 
    \item Reasonable fit: elements are compatible, though not deeply integrated
    \item Moderate coherence: combination makes general sense with limited justification
    \item 
    \item Coherent and justified: combination makes sense and is literature-informed
    \item 
    \item Highly coherent: seamless integration of ideas with strong literature basis
\end{enumerate}

\textbf{\textcolor{purple}{Explanation}}: You should also provide a rationale for your score. (Your rationale should be at least 2-3 sentences.)
\textit{Hint: If the fit between components is strong, note which elements are well connected and clearly defined for implementation (e.g., problem-task pairing, method-metric match). If weak, specify which parts feel vague, disconnected, or hard to execute.}

\textbf{\textcolor{purple}{9. Specificity of Proposed Method}}: How clearly does the proposed method present a testable research goal or hypothesis? To what extent is it sufficiently detailed to be operationalized in a way that aligns with prior literature or accepted practices?

\begin{enumerate}
    \item Extremely unclear: the method is explained in an extremely vague or ambiguous manner, making it impossible to understand or replicate the approach without additional information or clarification.
    \item 
    \item Unclear: the method is described with some detail, but significant gaps in explanation or logic leave the reader with considerable confusion and uncertainty about how to apply or replicate the approach.
    \item 
    \item Somewhat clear: method is described with sufficient detail to understand the basic approach, but important elements remain vague or underdeveloped
    \item Moderately clear: method is described with sufficient detail to understand the basic approach, but lacks the precision or specificity needed to fully replicate or grasp the nuances of the methodology without further guidance.
    \item 
    \item Clear and testable: method is clearly and precisely described, with most details provided to allow for replication and comprehension, though minor areas may benefit from further clarification or elaboration.
    \item 
    \item Highly clear and specific: method is articulated in an exceptionally clear, precise, and detailed manner, enabling straightforward replication and thorough understanding of the approach with no ambiguities
\end{enumerate}

\textbf{\textcolor{purple}{Explanation}}: You should also provide a rationale for your score. (Your rationale should be at least 2-3 sentences.)

\textbf{\textcolor{purple}{10. Overall Score}}: Overall score:  Apart from the above, you should also give an overall score for the idea on a scale of 1 - 10 as defined below (Major AI conferences in the descriptions below refer to top-tier NLP/AI conferences such as *ACL, COLM, NeurIPS, ICLR, and ICML.):

\begin{enumerate}
    \item Critically flawed, trivial, or wrong, would be a waste of students’ time to work on it
    \item Strong rejection for major AI conferences
    \item Clear rejection for major AI conferences
    \item Ok but not good enough, rejection for major AI conferences
    \item Decent idea but has some weaknesses or not exciting enough, marginally below the acceptance threshold of major AI conferences
    \item Marginally above the acceptance threshold of major AI conferences
    \item Good idea, would be accepted by major AI conferences
    \item Top 50\% of all published ideas on this topic at major AI conferences, clear accept
    \item Top 15\% of all published ideas on this topic at major AI conferences, strong accept
    \item Top 5\% of all published ideas on this topic at major AI conferences, will be a seminal paper
\end{enumerate}

\textbf{\textcolor{purple}{Overall Rationale}}: You should also provide a rationale for your overall score. (Your rationale should be at least 2-3 sentences.)
\textit{Hint: This is just an idea. Please evaluate its potential — assuming it is properly fleshed out, implemented, and empirically validated, would it be acceptable at a future major AI conference? If the idea is too vague to envision as a strong paper, it should be rated lower.}

\textbf{\textcolor{purple}{11. Confidence}}: Additionally, we ask for your confidence in your review on a scale of 1 to 5 defined as following:

\begin{enumerate}
    \item Your evaluation is an educated guess
    \item You are willing to defend the evaluation, but it is quite likely that you did not understand central parts of the paper
    \item You are fairly confident that the evaluation is correct
    \item You are confident but not absolutely certain that the evaluation is correct
    \item You are absolutely certain that the evaluation is correct and very familiar with the relevant literature
\end{enumerate}

\textbf{\textcolor{purple}{Time}}: How many minutes did you spend on this task?

\newpage

\subsection{Statistical Tests for Human Evaluation}
\label{app:human_eval_stats}

\begin{table}[t]
\centering
\scriptsize
\begin{tabular}{lcccccccc}
\toprule
 & \multicolumn{4}{c}{\textbf{Baseline}} & \multicolumn{4}{c}{\textbf{HARPA}} \\
\cmidrule(lr){2-5} \cmidrule(lr){6-9}
\textbf{Dimension} & Mean & Std & Min--Max & Median & Mean & Std & Min--Max & Median \\
\midrule
Coherence     & 7.20 & 1.65 & 3--10 & 8.0 & 7.08 & 1.65 & 1--10 & 8.0 \\
Confidence    & 4.33 & 0.66 & 3--5  & 4.0 & 4.38 & 0.67 & 3--5  & 4.0 \\
Effectiveness & 6.78 & 1.66 & 3--10 & 7.0 & 6.65 & 1.69 & 1--9  & 7.0 \\
Excitement    & 6.45 & 1.54 & 3--8  & 7.0 & 6.25 & 1.48 & 3--8  & 6.0 \\
Familiarity   & 3.93 & 0.94 & 2--5  & 4.0 & 4.05 & 0.93 & 2--5  & 4.0 \\
Feasibility   & 5.50 & 1.72 & 3--9  & 5.0 & 6.28 & 2.08 & 1--10 & 6.0 \\
Grounding     & 5.98 & 1.94 & 2--10 & 6.0 & 6.83 & 1.47 & 3--10 & 7.0 \\
Motivation    & 6.13 & 1.64 & 3--9  & 6.0 & 6.45 & 1.43 & 3--9  & 7.0 \\
Novelty       & 6.43 & 1.32 & 3--8  & 6.0 & 5.98 & 1.33 & 2--8  & 6.0 \\
Overall       & 6.20 & 1.71 & 3--8  & 7.0 & 6.43 & 1.69 & 2--9  & 7.0 \\
Specificity   & 7.00 & 1.88 & 2--10 & 8.0 & 7.28 & 1.78 & 1--10 & 8.0 \\
\bottomrule
\end{tabular}
\caption{Expert ratings across 11 dimensions. Values report mean, std, min--max, and median (10-point Likert scale, higher is better). $n=40$ proposals per system.}
\label{tab:expert_scores_full}
\end{table}

For each dimension, we computed paired differences between HARPA and the baseline on expert ratings. Statistical significance was assessed using bootstrap resampling (10,000 iterations) and Wilcoxon signed-rank tests. We report bootstrap as our primary test, since it makes no distributional assumptions and is appropriate for small sample sizes and ordinal scores. 
Table~\ref{tab:human_eval_stats} shows the mean differences and $p$-values.

\begin{table}[ht]
\centering
\begin{tabular}{lccccc}
\toprule
Dimension & MeanDiff & Bootstrap\_p & Boot* & Wilcoxon\_p & Wilcoxon* \\
\midrule
coherence\_score     & -0.125 & 0.666 &    & 0.806 &    \\
effectiveness\_score & -0.125 & 0.663 &    & 0.716 &    \\
excitement\_score    & -0.200 & 0.753 &    & 0.435 &    \\
familiarity\_score   &  0.125 & 0.210 &    & 0.394 &    \\
confidence\_score    &  0.050 & 0.360 &    & 0.660 &    \\
feasibility\_score   &  0.775 & 0.017 & *  & 0.016 & *  \\
grounding\_score     &  0.850 & 0.002 & ** & 0.017 & *  \\
motivation\_score    &  0.325 & 0.163 &    & 0.286 &    \\
novelty\_score       & -0.450 & 0.937 &    & 0.107 &    \\
overall\_score       &  0.225 & 0.275 &    & 0.598 &    \\
specificity\_score   &  0.275 & 0.168 &    & 0.430 &    \\
\bottomrule
\end{tabular}
\caption{Mean differences (HARPA – baseline) with significance tests. 
Stars indicate significance (* $p<0.05$, ** $p<0.01$). 
Bootstrap resampling is our primary test.}
\label{tab:human_eval_stats}
\end{table}

\section{Full examples of expert review and proposals}\label{app:expert_examples}

Table~\ref{tab:expert_example} shows two representative pairs of hypotheses (HARPA vs.\ baseline), along with expert assessment across all evaluation dimensions. Each row corresponds to one proposal. Complete dataset generated and assessed for human evaluation are available in the supplementary files.

\begin{table}[h!]
\centering
\scriptsize
\begin{tabular}{p{0.15\linewidth} p{0.8\linewidth}}
\toprule
\textbf{System} & \textbf{Hypothesis/Problem Statement, Ratings, and Justification} \\
\midrule
HARPA & 
\textit{``Integrating stochastic softmax tricks with control variates will significantly improve convergence speed and stability in spanning tree optimization problems compared to using stochastic softmax tricks alone.''} \\
& \textbf{Novelty = 6} (``The following paper is a neurips 2020 paper that has spanning tree optimization as an application: ""Gradient Estimation with Stochastic Softmax Tricks
"". It mainly uses softmax trick for some discrete problems such as spanning tree optimization. The main novelty comes to add the control variates into the loop.") \\
& \textbf{Feasibility = 7} (``Using the softmax trick allows backpropagation/gradient estimation, it is a well known trick and the implementation is not so complicated, although thee control variates is not so clear how would be implemented (with examples) ") \\
&\textbf{ Grounding = 9} (``It is very grounded on the listed refereneces, mostly similar to "Gradient Estimation with Stochastic Softmax Tricks (2020)". The control variates ideas, although slightly mentioned in the same paper, are more frequent described at "A generalized approximate control variate framework for multifidelity uncertainty quantification (2018)" ") \\
& \textbf{Specificity = 6 } (``I feel the control variates part is not so clear how it would be implemented. What is the additional variable that is correlated to the target? what would be the target in that case? I can see the motivation for that, but it is not so clear how it would be implemented. Examples would be appreciated. ") \\
&\textbf{ Coherence = 8} (``It is very clear that softmax trick is for gradient estimation and control variates is where the novelty is, to improve efficiency. So I see it is coherent. ") \\
& \textbf{Motivation = 6} (``The motivation only comes from adding control variates to speedup convergence to the already existing methods using "stochastic softmax tricks". I can't see anything else regarding motivation.  ") \\
& \textbf{Excitement = 5} (``"I would say it is not exciting due to the lack of novelty (compared to the given neurips paper in 2020). The experiments are also vanilla (mostly ablation studies). The experiment overview is basically removing the control variates and the softmax to compare with the method with both. ") \\
& \textbf{Effectiveness = 6} (``It is very clear that softmax trick is for gradient estimation and control variates is where the novelty is, to improve efficiency. So I see it is coherent. ") \\
& \textbf{Overall = 6 }(``The motivation of the proposed method is clear, increase efficiency when bringing the control variates into the loop. However, details on how to incorporate the control variate ideas, which is the core of the novelty, are not so clear.  ") \\
& \textbf{Confidence = 4} \\
\midrule

Baseline & 
\textit{``Gradient-based optimization in deep learning often suffers from instability and slow convergence, especially in complex decision-making pipelines where gradients can become extremely large or vanishingly small. This issue can lead to poor model performance, slow training, and difficulties in fine-tuning models for specific tasks.''} \\
& \textbf{Novelty = 5} (``From one perspective, I don't score high the novelty regarding this proposal since this would depend on some detais that are not explicitly mentioned. Example: "During each update, we compute a clipping threshold as a function of these statistics". How exactly those statistics will be used would impact on the novelty. For example, Adam optimizer also use statistics for updating gradients. However, the overall method has its novely characteristics by combining the statistics with stochastic perturbations.") \\
& \textbf{Feasibility = 4} (``The proposed method looks feasible. The problem is that it lacks details. Everything related to the method is summarized in 2 or 3 lines in the "Proposed methods" without any mathematical language. How would we smooth the clipping function, for example?")\\
&\textbf{ Grounding = 2} (``The proposal reference list is not linked to the proposed methods. Soe papers I am really aware of have nothing to do with the method proposed. For example, "Differentiation of Blackbox Combinatorial Solvers (2019)" is specifically about imitation learning of combinatorial labels, there is no novelty towards unconstrained optimizers.") \\
& \textbf{Specificity = 2} (``As I mentioned before, the details are the problem in this proposal. There is no details of the proposed methods, and therefore the specificity is extremely unclear.")\\
& \textbf{Coherence = 6} (``The proposed method is very weak and lack of important details. The experimental plan looks all correct, but they are not so important with respect to the method itself. e, it is obvious from the Proposed metod that it is a new optimizer, and then the step 1 of experimental plan it just repeat the steps without detailing it.")\\
& \textbf{Motivation = 6 }(``The section of motivation makes sense, although not grounded in the list of papers.") \\
& \textbf{Excitement = 7} (``It is definitely exciting in a sense that the impact of this type of research is extremely high, since it can substitute, for example, specific pytorch optimizers that are widely used (for example, Adam) and sometimes suffer from convergences difficulties depending on the architecture used (for example, RNNs).") \\
& \textbf{Effectiveness = 3} (``The main problem of this research proposal is the lack of details in the proposed method subsection. I still don't see how the parts that matter would be really implemented, such as the clipping part. And what is end-to-end? What is optimized for the adaptation?") \\
&\textbf{ Overall = 3} (``The main problem of this proposal is the lack of details. The method itself, in a high-level idea, makes sense. But the "how" is extremely unclear. There is no depth of the methodology. The ideas ends up in the high-level buzzwords.") \\
& \textbf{Confidence = 4} \\
\bottomrule
\end{tabular}
\caption{Representative HARPA vs.\ baseline hypotheses with expert assessment.}
\label{tab:expert_example}
\end{table}

\newpage

We include an example of full proposal evaluated by experts and generated by HARPA and by the baseline ideator.

\subsection*{HARPA Proposal}\label{app:harpa_example}
\includepdf[pages=-,scale=0.9]{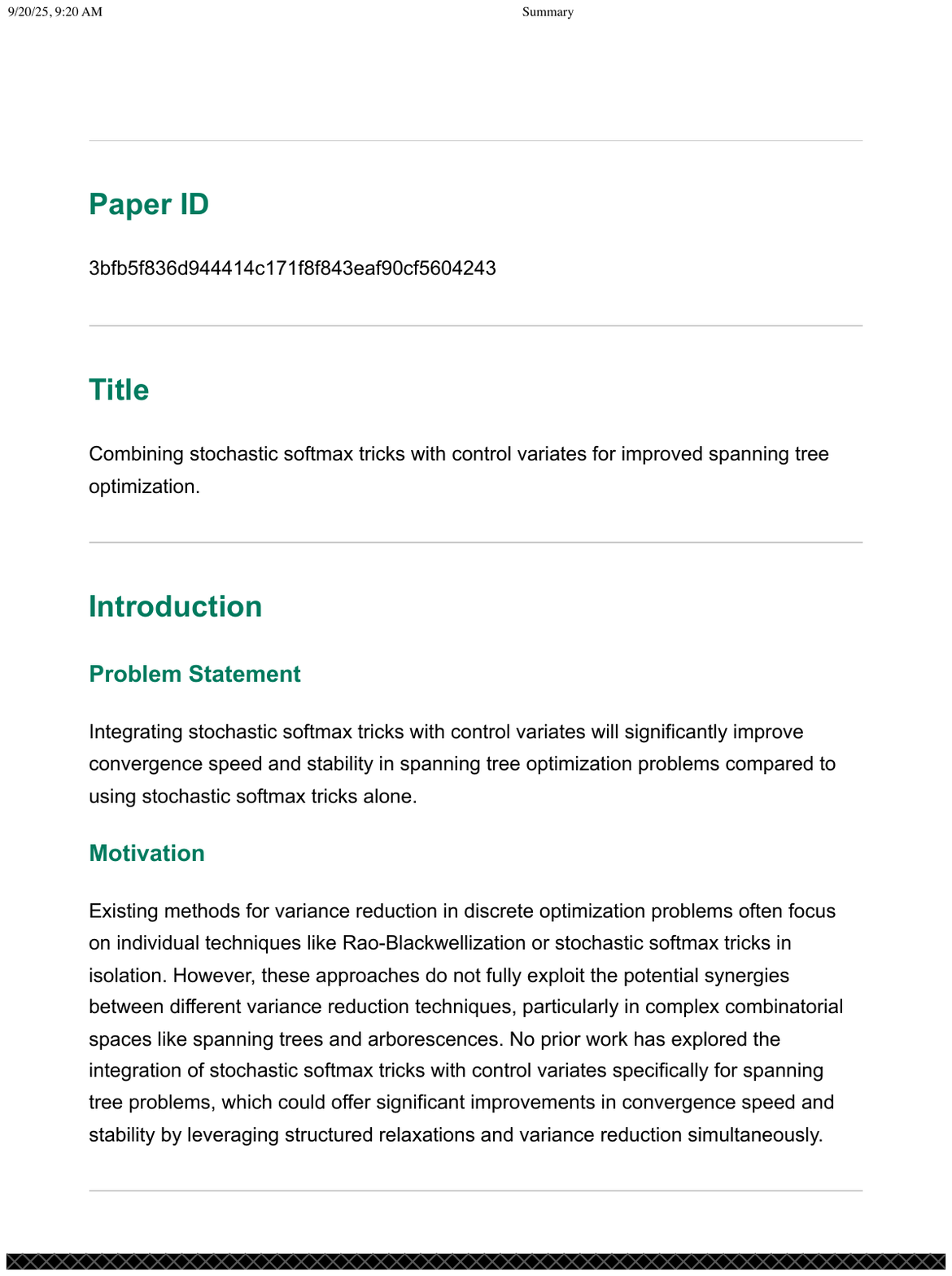}

\subsection*{Baseline Proposal}\label{app:baseline_example}
\includepdf[pages=-,scale=0.9]{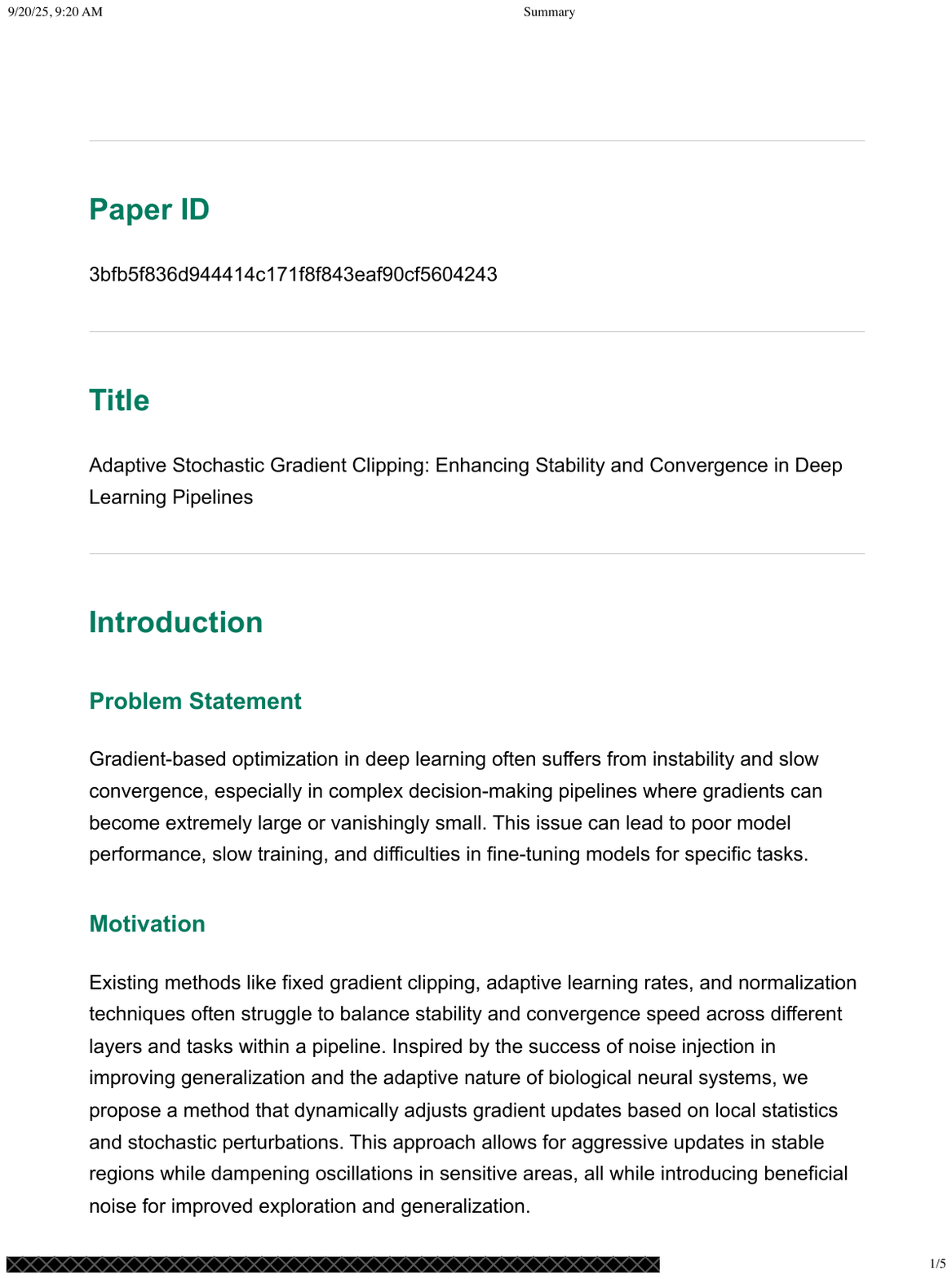}

%% file: appendix/harpa_scorer.tex
\section{HARPA-Scorer}\label{appendix:harpa_scorer_example}

\subsection{HARPA-Scorer: Additional Experiment Details}\label{app:implementation}
\paragraph{Implementation details.} 
We follow almost the same hyperparameters as the RM-R1 pipeline~\citep{chen2025rm}, consisting of two stages: (i) reasoning distillation from oracle rubric-style traces, and (ii) RLVR fine-tuning on execution-derived preference pairs. The backbone is \texttt{Qwen2.5-7B-Instruct}, trained with \texttt{openrlhf} and DeepSpeed using full fine-tuning. 

\textbf{Distillation Stage.} 
We fine-tuned on 3,459 rubric-aligned preference pairs (Section~\ref{sec:agent-exp}), split into 2,595 train, 452 validation, and 412 test. Training used a global batch size of 4 (micro-batch size 1), 
maximum sequence length 12{,}288, and Adam optimizer with offloading at a learning rate of $5\times 10^{-6}$. We trained for 1 epoch in \texttt{bfloat16} precision with ZeRO stage-2 optimization, gradient checkpointing, FlashAttention, and sample packing. Training was performed on 4$\times$ NVIDIA A100-SXM4-80GB.

\textbf{RLVR Stage.} 
We further optimized the distilled model with RLVR on execution-derived success/failure pairs, using the \texttt{verl} PPO trainer. Training was performed on 8$\times$NVIDIA A100-SXM4-80GB GPUs. We set the rollout batch size to 64, PPO mini-batch size to 16, and micro-batch size per GPU to 4. The learning rate was $1\times 10^{-6}$ with constant warmup. Maximum input and output lengths were both 8{,}192 tokens. KL regularization was applied with a coefficient of $10^{-3}$ and a clip ratio of 0.2, with entropy coefficient set to 0. GPU memory utilization was capped at 0.5 to prevent OOM issues. Sampling used temperature 1.0 and top-p 1.0. Training ran for a single epoch, with checkpoints saved every 1,000 steps.

\showtracebox{app:reasoning_trace}
{Example reasoning trace}
{prompts/reasoning_trace_example.txt}
{Example of rubric-style reasoning trace given the proposal pairs and their execution metadata from an oracle model}

\showjsonbox{app:cs_log_example}
{Example extracted execution trace}
{prompts/part_of_CS_logs_example.json}
{json}
{Example JSON snippet showing execution-derived factors from CodeScientist logs}

\section{HARPA Scorer Prompts}\label{appendix:harpa_rm_prompts}
In this section, we include all the prompts used for different tasks within the scorer \texttt{harpa-rm} pipeline.

\showscorerbox{app:trace_generation}
{System prompt for oracle reasoning trace generation}
{prompts/trace_generation.txt}
{Generate rubric-style reasoning trace given the proposal pairs and their execution metadata from an oracle model}

\showscorerbox{app:sft_data_generation}
{System prompt for SFT dataset generation}
{prompts/sft_instruction.txt}
{Generate SFT dataset with reasoning trace generation instruction without metadata information from \textsc{CodeScientist} execution traces.}

\showscorerbox{app:source_paper_topic}
{Topic extraction from abstracts}
{prompts/source_paper_topic_extraction.txt}
{Extract source paper abstract topics to construct preference pairs across same topics}

\showtracebox{app:baseline_reasoning_trace}
{Baseline predicted reasoning trace}
{prompts/baseline_reasoning_trace.txt}
{Example of rubric-style reasoning trace generated by the baseline LLM scorer. Note that while formatted as a rubric, the reasoning is unstructured, fails to follow instructions, and does not provide actionable justifications, making it unsuitable for guiding proposal refinement.}

%% file: appendix/harpa_stages.tex
\clearpage
%\onecolumn

\section{HARPA Prompts}\label{appendix:prompts}
In this section, we include all the prompts used for different tasks within the HARPA pipeline.

\showjsonbox{app:variable_value_example}
{Example part of HARPA artefact illustrating key variable and value space}
{prompts/kv_examples.json}
{json}
{Example JSON snippet showing some key variables and values with detailed information extracted by the HARPA proposal generator}

\showmintedbox{app:preliminary_hyp}
{Generate preliminary hypothesis with rationale}
{prompts/initialGenFromChain.txt}
{Generate preliminary hypothesis with rationale after analyzing trends from temporal reasoning paper chains}

\showmintedbox{app:generalize_hypothesis}
{Generalize Hypothesis for Literature Search}
{prompts/claimGeneralize.txt}
{Generalized $H$ to progressive 4 levels of claims used for literature search}

\showmintedbox{app:SocraticQA_gen}
{Generate hypothesis specific questions}
{prompts/socraticQA.txt}
{Generate atleast 20 questions to refine the preliminary hypothesis $H$ to $H^{'}$}

\showmintedbox{app:refined_hypothesis}
{Refine Hypothesis based on Socratic QA}
{prompts/refineQA_hyp.txt}
{Refine the preliminary hypothesis $H$ to $H^{'}$ by answering Socratic questions and making it more specific}

\showmintedbox{app:variable_extraction}
{Key Concepts Extraction}
{prompts/key_var.txt}
{Extraction of key variables or concepts}

\showmintedbox{app:variable_space}
{Exploring Variable Space}
{prompts/var_space.txt}
{Exploring Variable Value Space given the set of key variables or concepts}

\showmintedbox{app:final_hypothesis}
{Final hypothesis and research proposal}
{prompts/final_hyp.txt}
{Converging to a novel and testable research hypothesis given the hypothesis space}